\documentclass{article}

\usepackage[dblblindworkshop, final]{neurips_2026}

\usepackage[utf8]{inputenc} %
\usepackage[T1]{fontenc}    %
\usepackage{hyperref}       %
\usepackage{url}            %
\usepackage{booktabs}       %
\usepackage{amsfonts}       %
\usepackage{nicefrac}       %
\usepackage{microtype}      %
\usepackage{xcolor}         %

\usepackage[pdftex]{graphicx}
\setcitestyle{numbers,square}
\usepackage{amsmath}
\usepackage{booktabs}
\usepackage{multirow}
\usepackage{longtable}
\usepackage{subcaption}

\hypersetup{pdfauthor={}, pdftitle={}}
\workshoptitle{NeurIPS 2026 Workshop on Neural Network Artifacts as a New Data Modality}
\title{Representational Geometry as a Diagnostic for Adversarial Robustness}

\author{%
  Brandon Ismalej \\
  Department of Computer Science \\
  California State University, Northridge \\
  Northridge, California, USA \\
  \texttt{brandon.ismalej.671@my.csun.edu} \\
  \And
  Alex Cloninger \\
  Department of Mathematics \\
  Hal\i c\i o\u{g}lu Data Science Institute \\
  University of California San Diego \\
  La Jolla, California, USA \\
  \texttt{acloninger@ucsd.edu} \\
  \And
  Gal Mishne \\
  Hal\i c\i o\u{g}lu Data Science Institute \\
  University of California San Diego \\
  La Jolla, California, USA \\
  \texttt{gmishne@ucsd.edu} \\
}

\begin{document}

\maketitle

\begin{abstract}
Adversarial training is a common technique for making neural networks more robust to small, intentionally crafted input perturbations, and robustness is commonly measured through accuracy under attack.
By exploring inter-network representations across different adversarial training objectives, we can analyze whether multiple networks organize their latent representations similarly.
We study Multilayer Perceptrons trained on MNIST under three adversarial training regimes and compare the resulting population training dynamics and geometry to those of a vanilla network. %
We use M-PHATE to embed data samples across training epochs and examine how the geometry of the data changes during training under adversarial objectives. 
On penultimate-layer representations we measure displacement between clean and perturbed points, distributional distance through maximum mean discrepancy, and nearest-neighbor recovery. 
Our visualizations and quantitative analyses show that adversarial training leads to structural changes in how networks organize internal data representations, and
 robust accuracy and representational stability can rank the same defenses in opposite ways.
These findings support a discussion of re-evaluating how practitioners approach the auditing of model robustness by treating latent representations and their training trajectories as diagnostic artifacts across sets of networks. 
\end{abstract}

\section{Introduction}
Robustness is most commonly reported as a single scalar, without identifying how the structure of internal representations is organized, or how that organization formed during training. 
Adversarial attacks introduce small, carefully crafted perturbations to data samples with the aim of inducing misclassifications~\cite{szegedyIntriguingPropertiesNeural2013, goodfellowExplainingHarnessingAdversarial2015d}.
Adversarial training is the standard method of defense for neural networks, where the network is trained on perturbed samples to learn to correctly classify attack samples~\cite{madryDeepLearningModels2019}.
However, multiple networks can achieve similar accuracy on perturbed samples, even if their internal representations differ structurally. 
As a scalar, accuracy lacks the power to report whether their representation geometries are truly different. 

Differences in internal representational geometry stem largely from adversarial training, where clean accuracy is consequently sacrificed for robustness~\cite{tsiprasRobustnessMayBe2019, zhangTheoreticallyPrincipledTradeoff2019} while also altering the input features that a network learns from~\cite{ilyasAdversarialExamplesAre2019}.
As a result, the representations learned from robust networks are structurally different from those of a standard network~\cite{tsiprasRobustnessMayBe2019, wangUnderstandingLearningRepresentations2018}. 
Given that robustness is largely a representational phenomenon, assessing defenses through a single behavioral metric strips researchers of the benefit of understanding the structural organization within a network that enables it to defend against an adversary.
While existing work leverages similarity metrics~\cite{kornblithSimilarityNeuralNetwork2019, raghuSVCCASingularVector2017}
to compare standard and robust representations, these methods are limited to static post-hoc analyses. 
As a result, they overlook the training dynamics in networks and how these shift under varying perturbation strengths across groups of networks.

In this work, we examine what adversarial training does inside a network, across regimes trained under varying adversarial objectives and attack strengths. 
Per-epoch traces and penultimate layer representations are computational artifacts produced through standard and adversarial training. 
Every network is examined with the same underlying test images, in their clean and perturbed forms, so the measured geometry reflects common data as arranged by each network, which allows networks themselves to be compared directly against each other. 
Across 45 networks under three adversarial training objectives and five perturbation budgets, we show that adversarial training reshapes representational geometry in ways that accuracy is unable to capture, with the largest reshaping belonging to the defense that scores best on accuracy.
This disagreement widens as the perturbation budget grows and holds across each geometric measure computed.

The main contributions of this work are as follows: 
\textbf{(1)} we measure the representation geometry of a population of adversarially trained neural networks uniformly across attack objectives and varying perturbation budgets; 
\textbf{(2)} we show that geometric analyses rank network robustness differently than standard evaluations under threat-model-matched attacks; 
\textbf{(3)} we show that geometry and accuracy capture different aspects of robustness, since a defended network can keep accuracy high while its representations drift substantially when attacked.

\section{Background}
\label{sect: background}

Let $f_\theta : \mathcal{X} \rightarrow \mathbb{R}^C$ be a classifier with 
loss $l(f_\theta(x), y)$.
An adversarial attack creates a misclassification-inducing perturbation $\delta$ constrained to $ \| \delta \|_p \le \epsilon$ where $\epsilon$ is the perturbation budget and $\| \cdot \|_p$ is the $l_p$ norm~\cite{szegedyIntriguingPropertiesNeural2013, goodfellowExplainingHarnessingAdversarial2015d}.
In the white-box case, 
attacks are constructed using the input gradient 
$g = \nabla_x l(f_\theta(x), y) $, which provides the direction of the steepest increase in loss.
We use single-step, white-box attacks that differ only in how $g$ is normalized before scaling by $\epsilon$.
The fast gradient sign method (FGSM)~\cite{goodfellowExplainingHarnessingAdversarial2015d} perturbs the input by $\epsilon \cdot \operatorname{sign}(g)$
moving every input value by $\epsilon$. 
The $l_2$- and $l_{\infty}$-bounded attacks compute $\epsilon \cdot g/\|g\|_2$ and $\epsilon \cdot g/\|g\|_{\infty}$, respectively.
All scalings are clamped to $[0,1]$.
A min-max objective is optimized during adversarial training, $ \operatorname{min}_{\theta} \mathbb{E}_{(x,y)} [ \underset{ \| \delta \|_p \le \epsilon}{\max} l(f_{\theta}(x +\delta),y) ]$, where the inner maximization is approximated by an attack, while perturbed inputs are generated. 
The maximum mean discrepancy (MMD)
\cite{grettonKernelTwoSampleTest2012} 
measures distances between feature distributions in a Reproducing Kernel Hilbert Space (RKHS), which is used as an attack-agnostic measure of global representational difference.
For the $n$ matched pairs, we compute the estimate
$\widehat{MMD}^2 = \frac{1}{n^2} \sum_{i,i'} k(x_i, x_{i'}) + \frac{1}{m^2} \sum_{j,j'}k(y_j, y_{j'}) - \frac{2}{nm} \sum_{i,j}k(x_i, y_j)$,
with Gaussian kernel $k$ %
whose bandwidth is chosen per configuration using the median heuristic (further details in Appendix \ref{app:mmd}).

M-PHATE~\cite{giganteVisualizingPHATENeural2019}, an extension of the PHATE dimensionality reduction method~\cite{moonVisualizingStructureTransitions2019}, tracks and visualizes representational dynamics across training epochs, embedding these high-dimensional network trajectories into a low-dimensional space while preserving local and global structures. 
Here we change the perspective and study how the geometry of data (i.e., samples), rather than hidden units, evolves throughout training. 
We describe the M-PHATE construction in Section \ref{methods_embeddings}.

\noindent \textbf{Related work:} 
The standard defense for neural networks is adversarial training which is known to transform representational structures by sacrificing clean accuracy~\cite{tsiprasRobustnessMayBe2019}, increasing reliance on robust features~\cite{ilyasAdversarialExamplesAre2019}, and reorganizing latent structures~\cite{engstromAdversarialRobustnessPrior2019}. 
To measure the structural differences between trained networks with a similarity measure, CKA~\cite{kornblithSimilarityNeuralNetwork2019} provides a static, pairwise comparison post-hoc by comparing centered kernel matrices.
SVCCA integrates singular value decomposition and canonical correlation analysis to analyze differences in representations~\cite{raghuSVCCASingularVector2017}. 
To analyze the inter- and intra-unit dynamical characteristics of a network throughout training, M-PHATE~\cite{giganteVisualizingPHATENeural2019} embeds the activation tensor as a single object in low-dimensional space. Setting dimensionality to two or three enables visualizing the dynamics. 
This places our work alongside an increasingly vital domain that treats trained networks themselves as data to be analyzed directly. 
These efforts include predicting model properties from weights~\cite{unterthinerPredictingNeuralNetwork2021, eilertsenClassifyingClassifierDissecting2020}, compiling datasets of model populations~\cite{schurholtModelZoosDataset}, and developing architectures that operate on weight spaces~\cite{navonEquivariantArchitecturesLearning2023}. 
While weights are not well-ordered, activations are, which makes matched-pair displacement between clean/perturbed input pairs tractable.

\section{Setup and methods}
\label{sect:methods}

\subsection{Models and training}

All experiments use MNIST~\cite{lecunMNIST1998} with a fixed sample of 1,000 training images and 2,000 test images. We compute geometric measurements on a fixed batch of 1,000 test images with indices consistent across all experiments, so only initialization and batch ordering vary across runs. 
The architecture is a three-layer fully connected MLP ($784 \rightarrow 128 \rightarrow  64 \rightarrow 10$) with ReLU activations. 
Let $h_1 \in \mathbb{R}^{128}$  and $h_2 \in \mathbb{R}^{64}$ be the two hidden layers, where all geometric measurements are calculated on the penultimate layer $h_2$. 
We use fully connected MLPs so that representations are free from pooling or convolutional structure, and the training objective directly contributes to the measured geometry rather than an architectural prior. 
For that same reason, we use single-step attacks,
though multi-step attack suites exist
\cite{croceReliableEvaluationAdversarial2020d}.
We train under four regimes: \textit{vanilla} on clean data, and three adversarial regimes applying single-step attacks on every batch at each epoch.
Each regime is trained across five perturbation budgets referred to as Extreme Low to Extreme High. 
Each network is trained on three seeds, resulting in a total population of 48 networks: 45 adversarially-trained and 3 vanilla. 
Training and attack details are expanded upon in Appendix \ref{app:training_details}.

\subsection{Trajectory and joint embeddings}
\label{methods_embeddings}

PHATE~\cite{moonVisualizingStructureTransitions2019} embeds high-dimensional data by representing the data as an affinity graph to capture local relationships, applying a diffusion operator to learn global structure, and reducing the dimensionality with MDS, resulting in a low-dimensional representation of the original data. 
M-PHATE~\cite{giganteVisualizingPHATENeural2019} builds on this, treating networks as dynamical systems where each epoch becomes a single slice in a multislice graph between hidden units.
To study how this local geometry evolves, %
M-PHATE connects a hidden unit to itself across all epochs through interslice edges in the multi-slice graph, allowing a diffusion process to capture temporal evolution in structural organization. While Gigante et al.~\cite{giganteVisualizingPHATENeural2019} track hidden units, here we track \emph{test samples} directly, making each embedded point a single image at a single training epoch. 
This corresponds to embedding the \textit{sample} $\times$ \textit{epoch} modes of the activation tensor, rather than the \textit{hidden unit} $\times$ \textit{epoch} modes (see Appendix).

Temporal analysis is especially critical in defense settings because standard and adversarial training may differ in the organization of the final representations they arrive at, and the path taken to form them.  
Final-epoch representations hold limited information power since they are unable to differentiate a network that reorganized early from one that drifted throughout, yet these two behaviors represent different learning processes. 
During training, the hidden activations of a test batch are recorded per-epoch to create a tensor of shape ($50 \text{ epochs } \times 1,000 \text{ samples } \times 128 \text{ units }$) for $h_1$ and shape ($50 \times 1,000 \times 64$) for $h_2$, so a single embedding displays the entire training cycle as one continuous object.
To compare clean and perturbed samples within a shared space, we stack penultimate layer representations of each test batch into a two-slice tensor of shape $(2 \times 1,000 \times 64)$ using the multi-slice graph construction of M-PHATE~\cite{giganteVisualizingPHATENeural2019}, with each clean point linked to its perturbed sample.
Since the interslice edge is a matched pair, each clean image remains linked to its perturbed counterpart with weight $\operatorname{exp}\{-(d_i/\sigma)^2\}$, where $d_i$ is the distance between the pair, and $\sigma$ is a global bandwidth set to their mean distance.
Following prior work using these embeddings directly as geometric representations~\cite{giganteVisualizingPHATENeural2019, xie-MMPHATE} all measurements in Section \ref{methods:geom} are computed in this space. 
We report results corresponding to using the raw $h_2$ activations in Appendix \ref{app:complete_results}.

\subsection{Geometric analysis}%
\label{methods:geom}
We feed clean and perturbed test batches through each network, extracting their penultimate representations $h_i ^{\text{clean}}$ and $h_i ^{\text{adv}}$, which are jointly embedded, as described in Section \ref{methods_embeddings}. All three quantities below are computed on the resulting 3-dimensional M-PHATE coordinates.
Our batches keep sample correspondence, so we measure matched pair quantities, described below.

\textbf{Normalized displacement} ($d_{\text{norm}}$) is the mean distance that the applied perturbation moves a point, expressed relative to the scale of that network's own representation space.
Since raw distances are not comparable across networks, we divide mean perturbation displacement by the within-network median to obtain a dimensionless quantity that measures how far a point moves after perturbation, relative to how far apart two arbitrary clean points lie. 
This yields a dimensionless, scale-invariant metric appropriate for cross-network comparison.

\textbf{Distributional divergence} is the measure of dissimilarity or difference between two distributions of data points. 
We use MMD with a Gaussian kernel to test whether the representations of clean and perturbed points are drawn from the same distribution. 
Since $\widehat{MMD}^2$ provides a scalar output, distributional difference is formally characterized using a permutation test with $N_{\text{perm}} = 1000$ permutations. Kernel parameters are fit per configuration. 

\textbf{Neighborhood recovery} requires no normalization, as we report the fraction of perturbed points whose clean counterpart appears within their $k=10$ nearest clean neighbors, making the measure invariant to any rescaling of the representation space.

\section{Representational drift from adversarial training}
\label{sect:results}

We report our main findings below: 
testing networks on common attacks aligns with geometric measurement ranking about robustness and similarity among points;
standard seen attack metrics rank training regimes opposite of geometric measurements;
accuracy-based and geometric metrics diverge in specific cases, showing 
that the two metrics are not interchangeable as measures of robustness. 

\subsection{Measuring robustness with accuracy can coincide with geometric stability}

\begin{figure}[htbp]
    \centering
    \captionsetup{skip=4pt}
    \captionsetup[sub]{skip=2pt, margin=0pt}
    \begin{subfigure}[b]{0.4\linewidth}
        \centering
        \includegraphics[width=\linewidth, trim=5 5 5 5, clip]{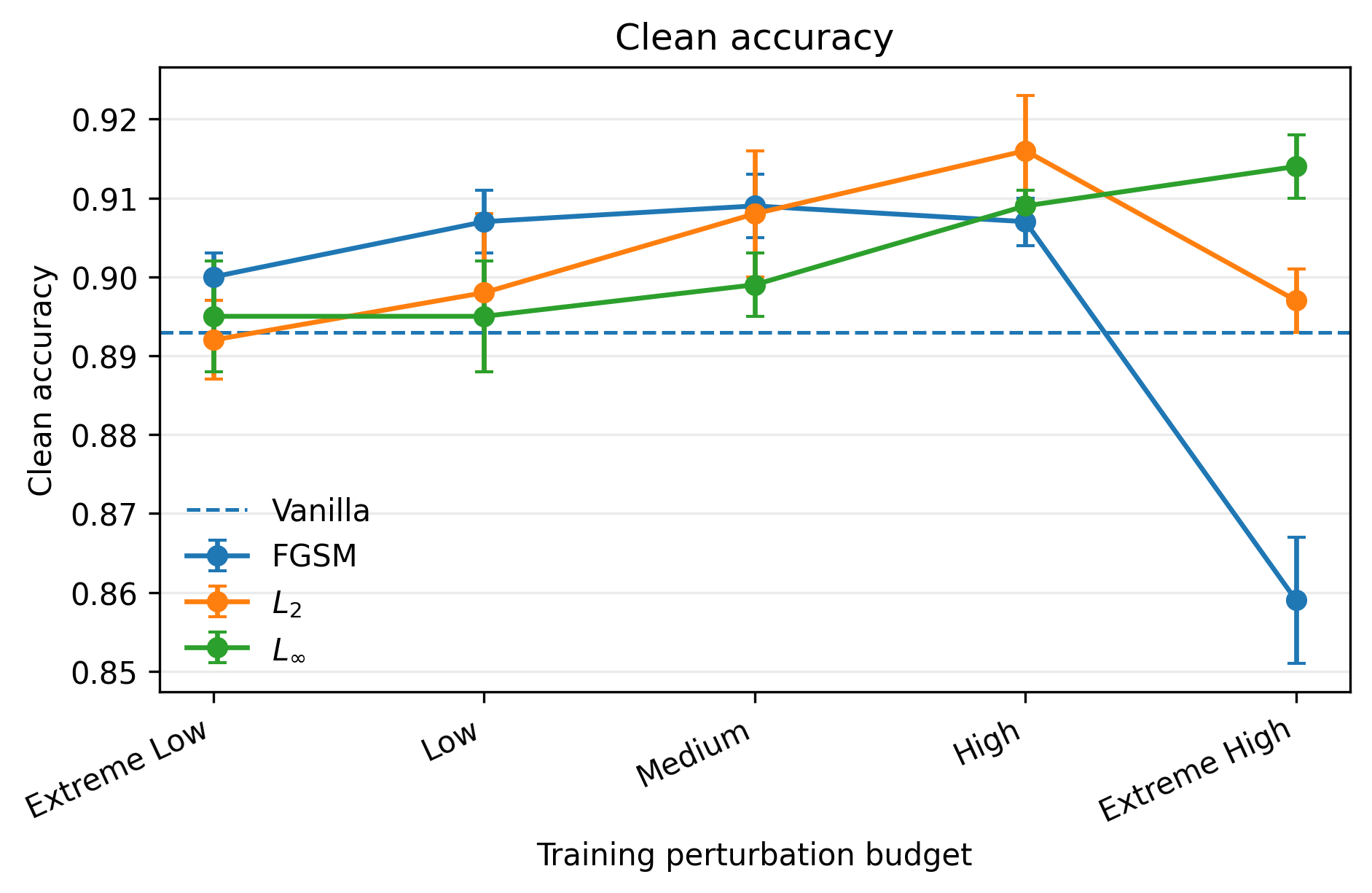}
        \label{fig:clean_acc}
    \end{subfigure}%
    \hspace{9mm}
    \begin{subfigure}[b]{0.4\linewidth}
        \centering
        \includegraphics[width=\linewidth, trim=5 5 5 5, clip]{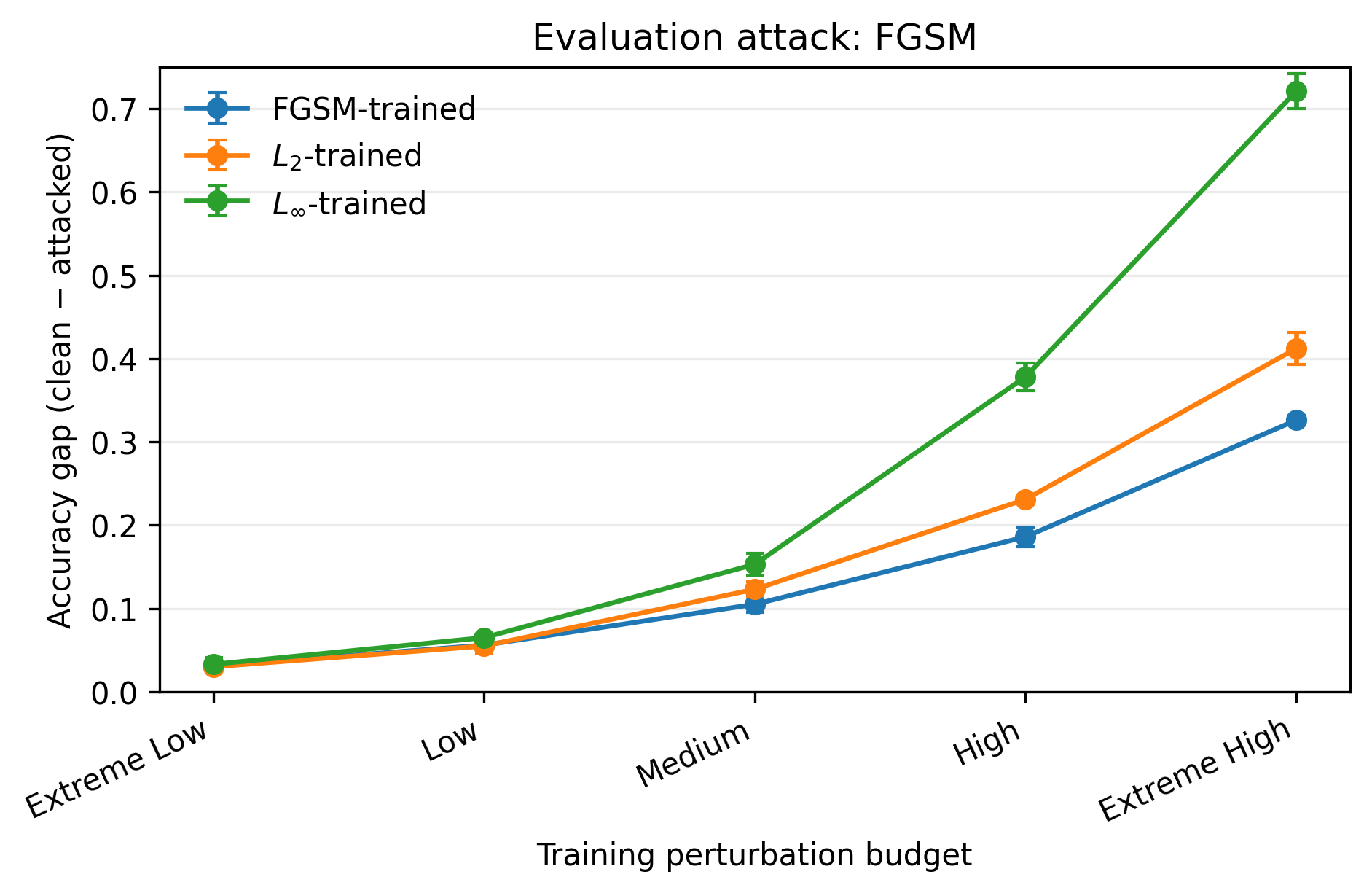}
        \label{fig:gap_fgsm}
    \end{subfigure}

    \vspace{-0.3em} %

    \begin{subfigure}[b]{0.4\linewidth}
        \centering
        \includegraphics[width=\linewidth, trim=5 5 5 5, clip]{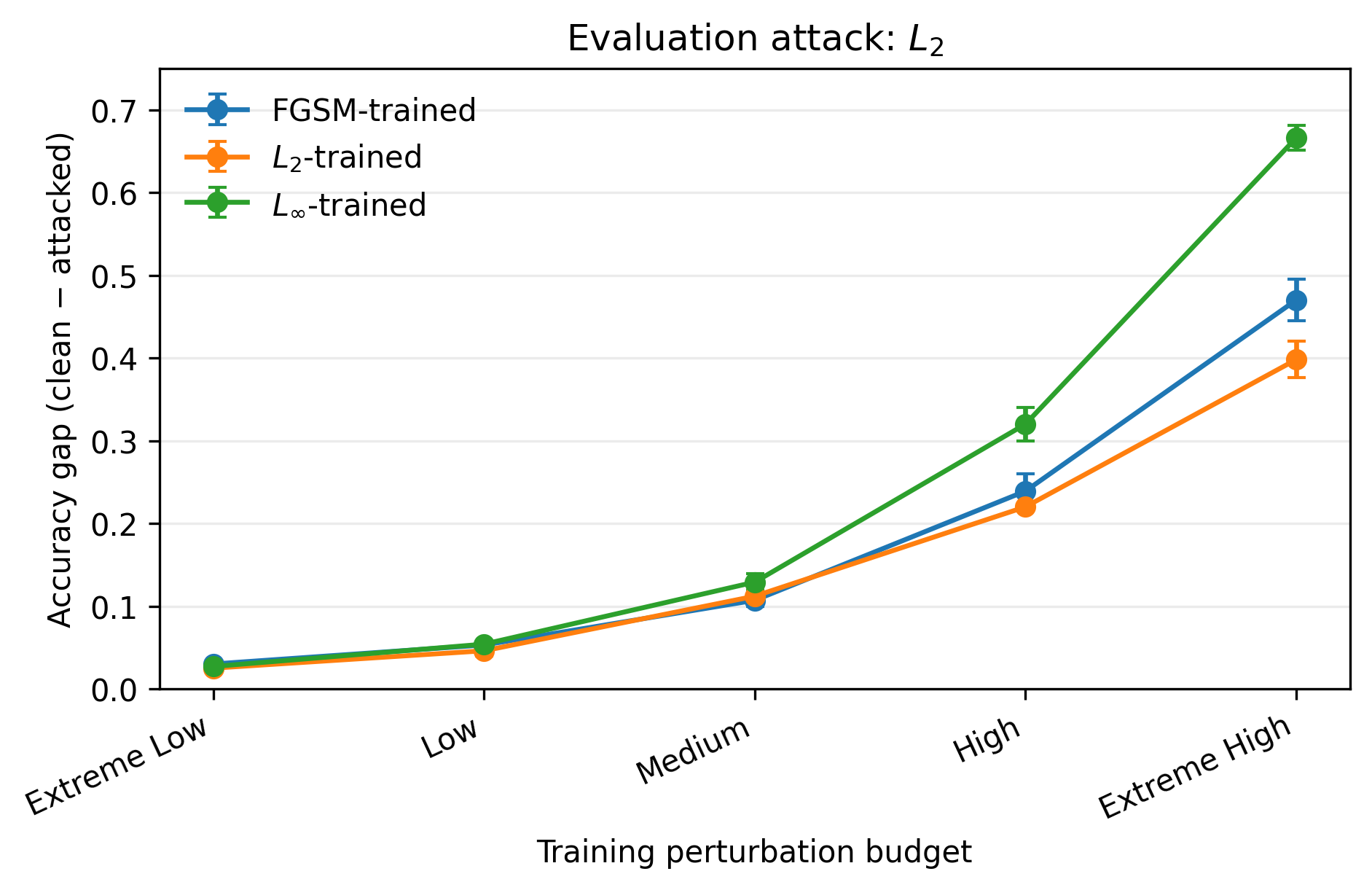}
        \label{fig:gap_l2}
    \end{subfigure}%
    \hspace{9mm}
    \begin{subfigure}[b]{0.4\linewidth}
        \centering
        \includegraphics[width=\linewidth, trim=5 5 5 5, clip]{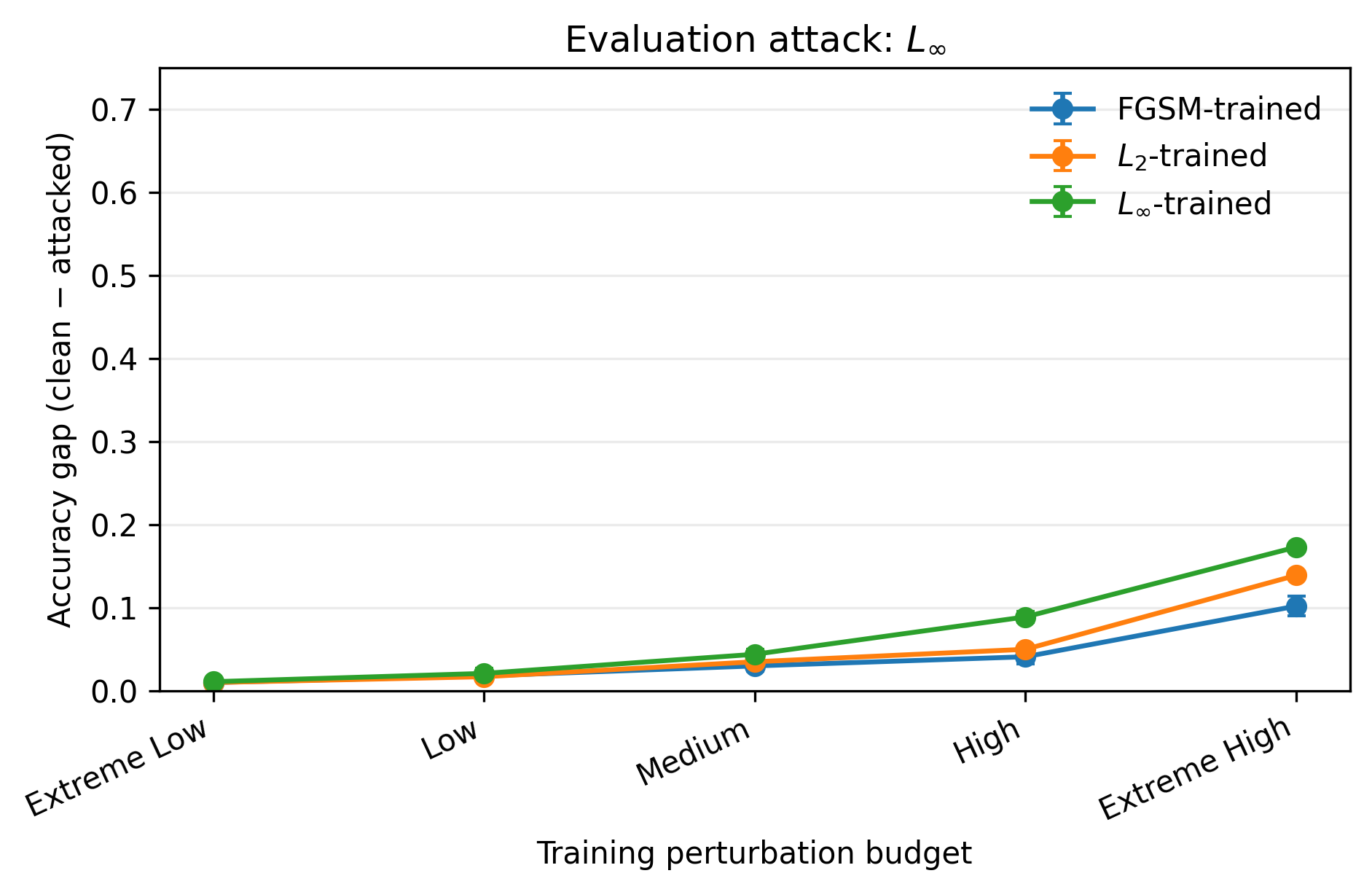}
        \label{fig:gap_linf}
    \end{subfigure}

    \caption{Clean accuracy and adversarial performance gaps for FGSM, $\ell_2$ and $\ell_\infty$ attacks across training perturbation budgets. A gap of $0$ means clean accuracy matches accuracy under attack and larger gaps mean a larger drop under attack.}
    \label{fig:accuracy_results}
\end{figure}

Figure \ref{fig:accuracy_results} reports clean accuracy of each trained network and its gap in accuracy against each attack along the same perturbation budget level: $\textbf{accuracy}_{\text{clean}} - \textbf{accuracy}_{\text{robust}}$.
Clean accuracy remains stable across regimes, except for the FGSM-trained network, which exhibits a known instability that can come from single-step adversarial training with large budgets~\cite{wongFastBetterFree2020}.

\begin{figure}[htbp]
  \centering
  \includegraphics[width=0.85\linewidth]{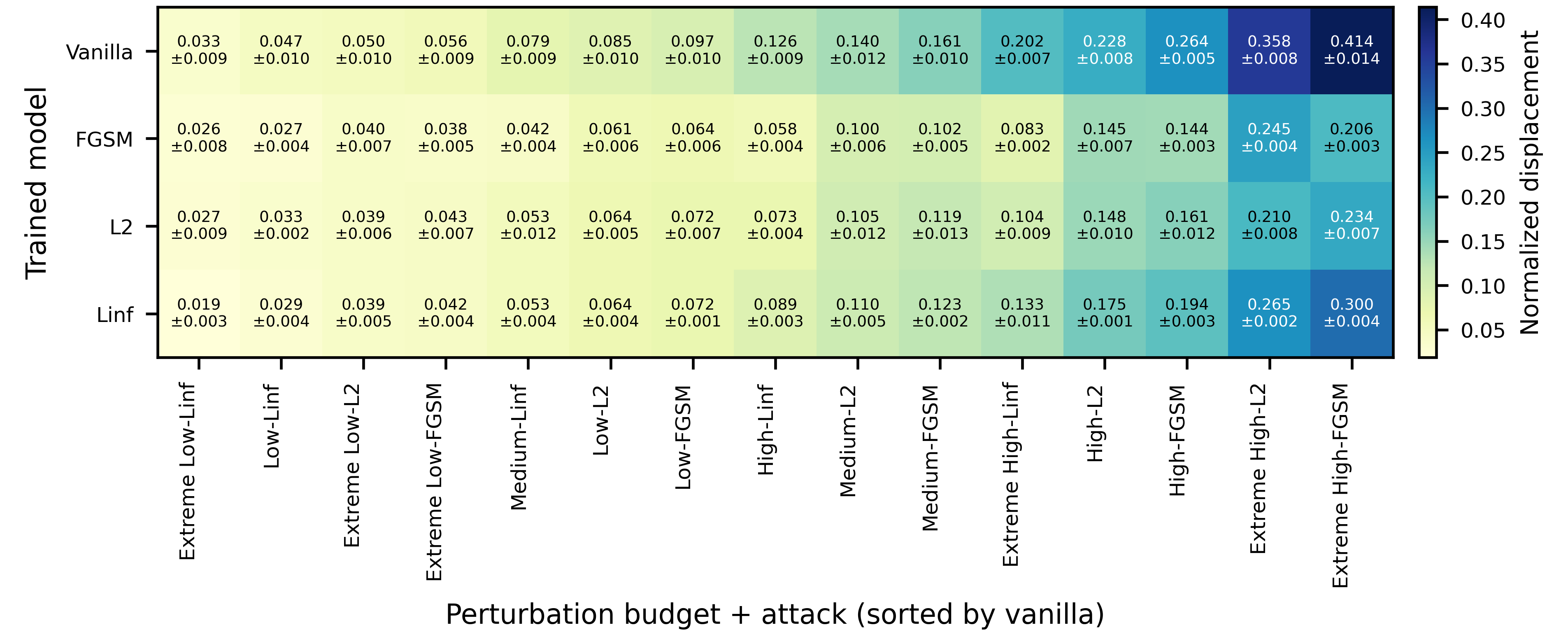}
  \caption{Normalized displacement between matched clean and perturbed representations in 3D M-PHATE space; larger values mean the perturbation moved representations further relative to the network's own scale. Rows: trained network. Columns: budget and attack used to perturb, ordered by ascending vanilla displacement. Each cell represents the network trained at the budget and perturbations of that attack at the same budget. Mean $\pm$ s.d., three seeds.}
  \label{fig:displacements}
\end{figure}

\begin{figure}[b]
  \centering
  \includegraphics[width=0.9\linewidth]{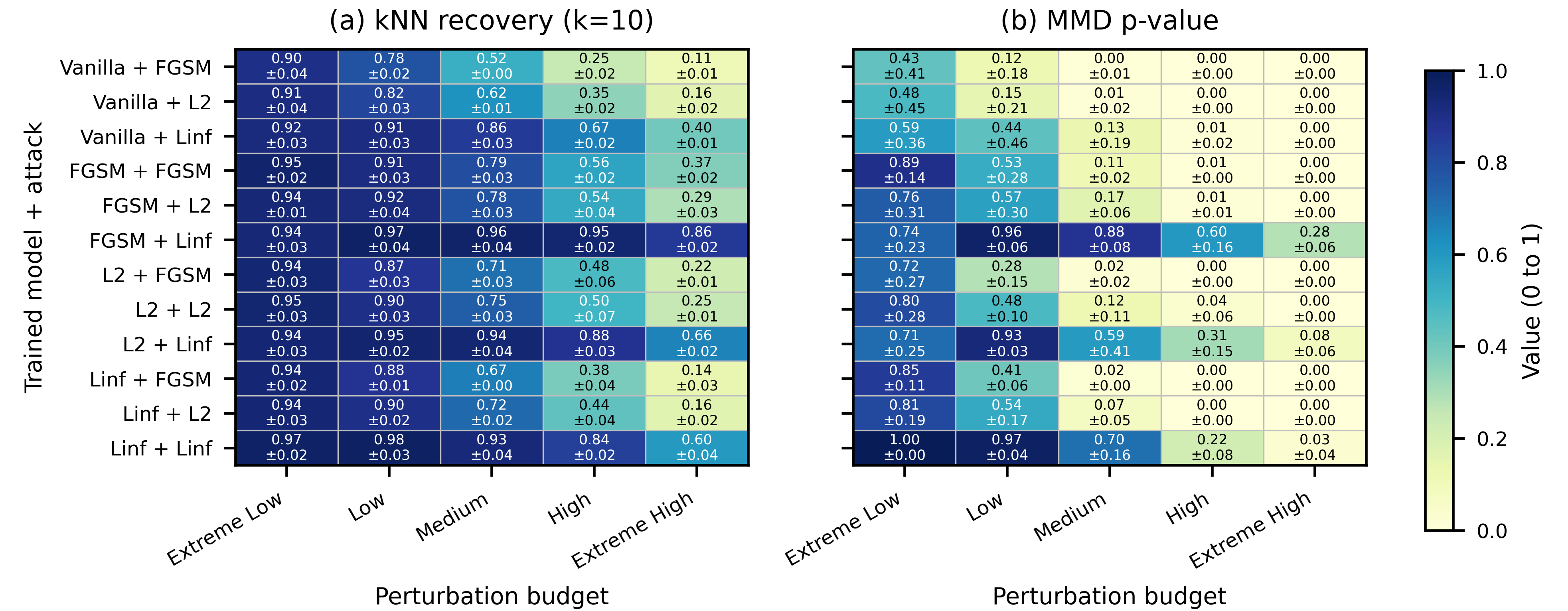}
  \caption{\textbf{(a)} Proportion of perturbed test points whose clean counterpart lies within its ten nearest neighbors in 3D PHATE space. \textbf{(b)} MMD permutation test $p$-values between clean and perturbed points in 3D PHATE space.  Recovery near $1$ and $p$ near $1$ both mean the perturbation left the representation essentially unchanged. For both panels, rows denote the trained network and applied evaluation attack, and columns denote the perturbation budget. %
  Mean $\pm$ s.d. across three seeds.}
  \label{fig:phate_heatmaps}
\end{figure}

\begin{figure}[t]
  \centering
  \includegraphics[width=0.8\linewidth]{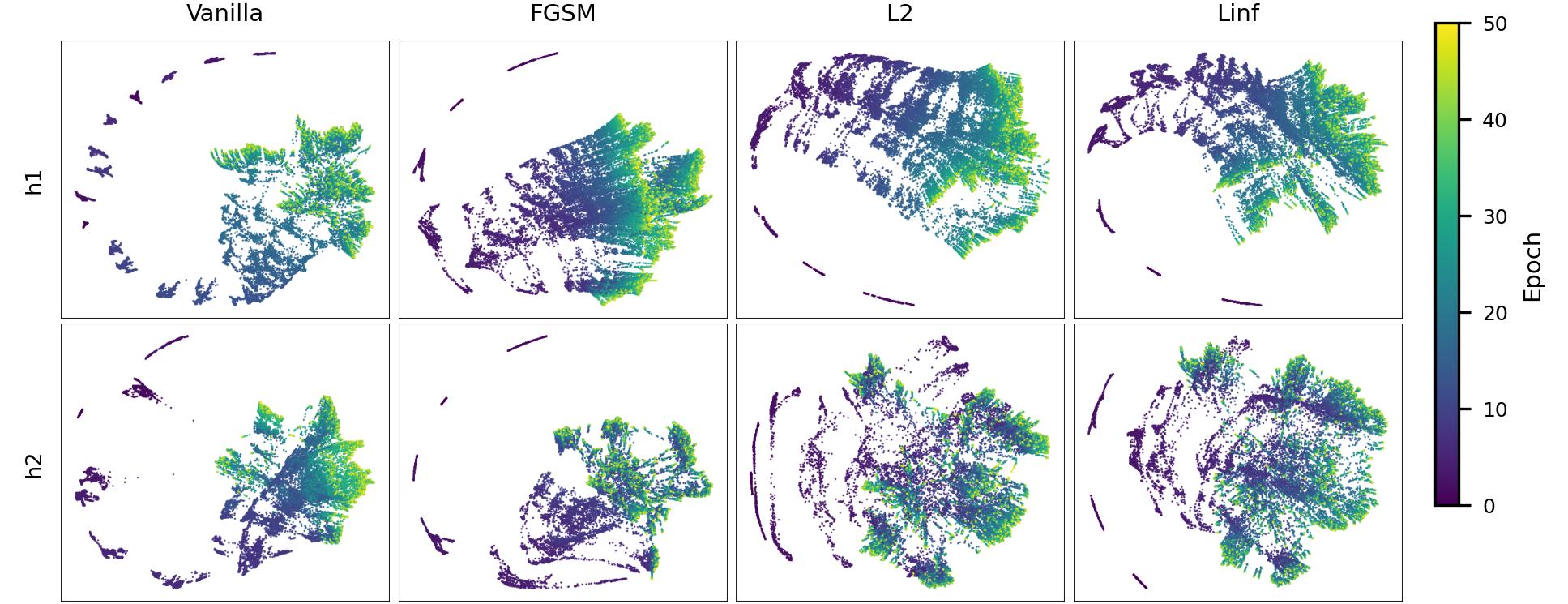}
  \caption{M-PHATE embeddings of $h_1$ (top) and $h_2$ (bottom) across all 50 training epochs at the High budget, one network per column. Each panel embeds the full trajectory of activations for the clean test batch, colored by epoch. \vspace{-0.6cm} }
  \label{fig:m-phate_medium_budget_h1_h2}
\end{figure}

Comparing each network against the attack it was trained on gives a standard measure of robustness, shown within Figure~\ref{fig:accuracy_results}
(Mean $\pm$ s.d. are shown in Appendix \ref{app:complete_results}).
A clear pattern arises here: $l_\infty$ exhibits the smallest gap between clean accuracy and robust accuracy, across its own attack. 
This suggests, by standard measures, that it ranks as the most robust network in this population.
Geometric measures suggest the opposite, ranking it last across the board. 
Across the three attack families, the $l_{\infty}$-trained network exhibits the largest mean displacement at Medium, High, and Extreme High, at 0.095, 0.153, and 0.233 (Figure \ref{fig:displacements}).
It also shows the lowest mean kNN recovery proportions, 0.77, 0.55, and 0.30 (Figure \ref{fig:phate_heatmaps} (a)), while the FGSM-trained network has the smallest displacement and highest recovery, with $l_2$ between them. 
Distributional discrepancy follows a similar pattern (Figure \ref{fig:phate_heatmaps} (b)), revealing that the defended network with the most appealing robustness number is the one that has moved its representations furthest from where they started: a defense can rearrange how a network represents its data with no guarantee that accuracy will reflect it. 

Figure \ref{fig:m-phate_medium_budget_h1_h2} shows the 2-dimensional M-PHATE embeddings for hidden-layer activations over 50 epochs for the four regimes at the High perturbation budget. 
Compared to the vanilla network dynamics, all adversarially trained networks show a clear pattern in either a single layer or both layers: learning stability begins to collapse at earlier epochs. 
The first hidden layer remains more stable through all 50 epochs than the second, where representation movement becomes difficult to track; the FGSM regime appears the least sporadic of the defended networks and closely resembles the vanilla dynamics.
All M-PHATE embeddings are in Appendix \ref{app:additional_figures}, where these behaviors vary across budgets, and give a qualitative view of clean versus perturbed representations.

\subsection{Ranking robust networks varies with evaluation attacks}

The vanilla network explains why the $l_{\infty}$-trained network appears most robust under threat-model-matched evaluation: under the same nominal $\epsilon$, an undefended network loses 0.232 accuracy under FGSM perturbation, but only 0.055 under $l_{\infty}$-bounded perturbation.
Standard evaluations introduce bias into comparison since different attack types have different optimization objectives. 
Geometric measurements complement this by evaluating each network against each type of attack equally, providing a natural architecture- and attack-agnostic benchmark, enabling practitioners to compare populations of networks without strict constraints or requirements. 

At Extreme High FGSM attack, the $l_\infty$-trained network provides little benefit in accuracy, reducing the vanilla gap of $0.856$ to $0.721$, where the FGSM-trained network reaches a far lower gap of $0.326$, yet it retains much of the geometric benefit, with displacement reduced from $0.414$ to $0.300$ against the FGSM-trained displacement of $0.206$.
Therefore, the measured geometric benefit of a defense is not proportional to the benefit that accuracy measures. 
We also note that no adversarially trained network has larger displacement than the vanilla network in any of the 45 cells, shown in Figure \ref{fig:displacements}.

\section{Discussion, limitations, and next steps}
\label{sect:discussion_limitations_next-steps}

Across a group of 48 networks spanning four regimes (three adversarial and one standard) and five perturbation budgets, 
rankings imposed by conventional robustness metrics are not necessarily in agreement with their geometric structures.
This nuanced relationship is attributed to the choice of evaluation attack used to quantify robustness. 
Scoring a defense against an attack it was trained on means that each network faces a test of different difficulty, whereas geometric measures enable cross-comparison across networks and test attacks without having to choose which attack represents a given network.
For instance, our undefended network loses 0.232 accuracy under FGSM attack, but only 0.055 under the $l_{\infty}$-bounded attack at the same nominal $\epsilon$, showing that
evaluating defenses at equivalent $\epsilon$-values subjects models to tests of unequal difficulty. 
Geometry does not replace accuracy, since accuracy only calculates whether or not a decision boundary was crossed, while displacement, MMD, and kNN recovery report how far representations shifted, whether their global distribution changed, and whether local neighborhoods survived.
Reporting both distinguishes a defense that stabilized a network's representations from one that only scored well against a favorable attack.
We acknowledge the results presented reflect
small MLPs on MNIST with single-step attacks, and are mainly derived from penultimate layer representations.
Perturbation budgets were chosen to follow convention within the domain literature, matching on a nominal $\epsilon$, though extended study could compute budgets matched on applied perturbation magnitude. 
In future work, we will extend this analysis to convolutional- and transformer-based architectures~\cite{heDeepResidualLearning2015, zagoruykoWideResidualNetworks2016, dosovitskiyImageWorth16x162021} on CIFAR-10 and ImageNet, where representational structure is richer and robustness claims carry practical weight for researchers.
Our results also indicate the geometry within neural network representations can be used for adversarial detection, using geometric shift as the basis for identifying perturbed inputs.

\newpage
\bibliographystyle{unsrtnat}
\bibliography{refs}

\newpage

\appendix

\section{Extended methodological details}
\label{app:background_details}

\subsection{Maximum mean discrepancy}
\label{app:mmd}
The maximum mean discrepancy (MMD)~\cite{grettonKernelTwoSampleTest2012} can be used to measure the distance between two sets of representations through their mean embeddings in a Reproducing Kernel Hilbert Space (RKHS). 
Let $X = \{x_i\}_{i=1}^n$ and $Y = \{y_j\}_{j=1}^m$ be the clean and perturbed representations, respectively, with $n=m=1000$ and pooled as $Z = [X,Y]$.
We use a Gaussian kernel
\[
k(u,v) = \operatorname{exp}(-\gamma \| u-v \|_2 ^2) \text{ with }
\gamma = \frac{1}{[\operatorname{median}_{i,j} \| z_i - z_j \|_2]^2 + \epsilon_0}
\]
with the bandwidth chosen using the median heuristic and $\epsilon_0 = 10^{-12}$.
The kernel matrix is computed once and held fixed. 
Using the kernel matrix $K \in \mathbb{R}^{(n+m) \times (n+m)}$ of $Z$ and the weight vector $w = (\frac{1}{n}, \dots, \frac{1}{n}, -\frac{1}{m}, \dots, -\frac{1}{m})^{\top}$,
the biased estimator of MMD is measured as:
\[
\widehat{MMD}^2 = w^{\top}Kw=  \frac{1}{n^2} \sum_{i,i'} k(x_i, x_{i'}) + \frac{1}{m^2} \sum_{j,j'}k(y_j, y_{j'}) - \frac{2}{nm} \sum_{i,j}k(x_i, y_j).
\]

Since this statistic is invariant to global rescaling of the representation space, comparison across networks is appropriate. 
To properly compare MMD scalars across networks, we assess statistical significance using a permutation test with $N = 1,000$ permutations.
Each permutation $\pi$ shuffles the pooled indices and recomputes $\widehat{MMD}_{\pi}^2 = w^{\top}_{\pi}Kw_{\pi}$.
The p-value is computed as
\[
p = \frac{1}{N} \sum_\pi 1[\widehat{MMD}_{\pi}^2 \ge \widehat{MMD}^2]
\]
which is reported as 0.000 when no permutation satisfies the inequality, indicating $p < 0.001$ rather than exact zero.

\subsection{Normalized displacement}
\label{ref:displacement}

Let $h_i ^{\text{clean}}$ and $h_i ^{\text{adv}}$ be the representations of the i-th clean example and its perturbed counterpart, in the space being measured. 
The normalized displacement is computed as
\[
d_{\text{norm}} = \frac{\frac{1}{n} \sum_{i=1}^n \| h_i^{\text{clean}} - h_i^{\text{adv}} \|_2}{\operatorname{median}_{i,j} \| h_i^{\text{clean}} - h_j^{\text{clean}} \|_2}.
\]

\subsection{kNN recovery}
\label{app:recovery}

For each perturbed representation, $h_i^{\text{adv}}$, let $\operatorname{\mathcal{N}_k}(h_i^{\text{adv}})$ be the index set of its k clean-representation nearest neighbors under Euclidean distance. 
The kNN recovery rate is
\[
R_k = \frac{1}{n} \sum_{i=1}^{n}1[i \in \operatorname{\mathcal{N}_k}(h_i^{\text{adv}})]
\]
where k=10 in this work.
$R_k = 1$ means every single perturbed point's clean counterpart is within its k nearest clean neighbors, and $R_k= 0$ means no perturbed point is within its k clean nearest neighbors.

\subsection{PHATE and M-PHATE }
\label{app:phate_and_mphate}
PHATE is a powerful dimensionality reduction technique that optimizes diffusion maps and aims to preserve both global and local structure for data visualization~\cite{moonVisualizingStructureTransitions2019}. 
It works by building a graph of data points, computing affinities to capture local relationships, and applying a diffusion operator to capture multiscale global structure. The resulting embedding maps potential distances into a low-dimensional space, maintaining the spatial organization of its high-dimensional counterpart. 
Built upon PHATE, M-PHATE extends this to time-variant high-dimensional data (i.e. network representations over epochs)~\cite{giganteVisualizingPHATENeural2019}. A tensor of shape $(\text{epochs} \times \text{samples} \times \text{units})$ is formed into a multislice graph where every epoch is a "slice", where interslice edges are connected to the same unit across epochs and intraslice edges connected within epochs. 
Applying diffusion on this combined graph creates an embedding where the representational trajectory of a network throughout training can be seen as a continuous structure, analogous to treating networks as dynamical systems whose temporal artifacts evolve along a low-dimensional trajectory. 
For further details, see Moon et al.~\cite{moonVisualizingStructureTransitions2019} and Gigante et al.~\cite{giganteVisualizingPHATENeural2019}.

\subsection{Attacks}
\label{app:attacks}
Let $g = \nabla_x l(f_{\theta}(x),y )$ be the input gradient of a single sample, and let $\Pi_{[0,1]}$ denote clamping to the valid pixel range. 
The three single step attacks used are
\[
x^{\mathrm{adv}}_{\mathrm{FGSM}} = \Pi_{[0,1]}\big(x + \varepsilon \cdot \mathrm{sign}(g)\big) 
\]
\[x^{\mathrm{adv}}_{l_2} = \Pi_{[0,1]}\Big(x + \varepsilon \cdot \tfrac{g}{\|g\|_2 + \epsilon_0}\Big)
\]
\[
x^{\mathrm{adv}}_{l_\infty}     = \Pi_{[0,1]}\Big(x + \varepsilon \cdot \tfrac{g}{\|g\|_\infty + \epsilon_0}\Big)\]
where norms are computed over the flattened image and $\epsilon_0 = 10^{-12}$.

\subsection{Network training and attack configurations}
\label{app:training_details}

Table \ref{tab:train-config} displays the details of training configurations used in this paper. This is held constant throughout all work in the paper, with the only variation being the adversarial training regime and perturbation budgets. 
Table \ref{tab:epsilons} displays perturbation budgets as fractions and decimals for each attack used. 
Table \ref{tab:magnitudes} shows the mean perturbation magnitudes for each attacked batch of 1000 images used for evaluation. 

\begin{table}[htbp]
\centering
\caption{Training configuration, identical for all networks and budgets.}
\label{tab:train-config}
\small
\resizebox{\textwidth}{!}{%
\begin{tabular}{ll}
\toprule
Item & Value \\
\midrule
Architecture & MLP, $784 \rightarrow 128 \rightarrow 64 \rightarrow 10$, ReLU \\
Optimizer & Adam \\
Learning rate $\eta$ & $0.001$ \\
Mini-batch size $B$ & 64 \\
Epochs & 50 \\
Loss & cross-entropy \\
\midrule
Dataset & MNIST \\
Training subsample & 1000 images drawn without replacement from the 60{,}000 training images \\
Subsample draw & \texttt{np.random.seed(42)}; \texttt{np.random.choice(60000, 1000, replace=False)} \\
Test subsample & first 2000 test images; evaluation uses the first 1000 of these \\
Subsample indices & fixed across all seeds and all budgets \\
\midrule
Seeds & 0, 1, 2 (\texttt{np.random.seed}, \texttt{torch.manual\_seed}) \\
\bottomrule
\end{tabular}}
\end{table}

\begin{table}[htbp]
\centering
\caption{Perturbation budgets. FGSM and $l_\infty$ use the same pixel-scale $\varepsilon$, given both as a fraction of 255 and as a decimal; $l_2$ uses a separate scale.}
\label{tab:epsilons}
\begin{tabular}{lccccc}
\toprule
Budget & \multicolumn{2}{c}{FGSM $\varepsilon$} & $l_2$ $\varepsilon$ & \multicolumn{2}{c}{$l_\infty$ $\varepsilon$} \\
\midrule
Extreme Low & 2/255 & 0.007843 & 0.125 & 2/255 & 0.007843 \\
Low & 4/255 & 0.015686 & 0.250 & 4/255 & 0.015686 \\
Medium & 8/255 & 0.031373 & 0.500 & 8/255 & 0.031373 \\
High & 16/255 & 0.062745 & 1.000 & 16/255 & 0.062745 \\
Extreme High & 32/255 & 0.125490 & 2.000 & 32/255 & 0.125490 \\
\bottomrule
\end{tabular}
\end{table}

\begin{table}[htbp]
\centering
\caption{Measured perturbation magnitudes, averaged over the 1000-image evaluation batch. $\delta = x_{\mathrm{adv}} - x$ is measured after clamping to $[0,1]$, so these are the perturbations actually applied rather than the nominal budgets.}
\label{tab:magnitudes}
\begin{tabular}{llcccc}
\toprule
Budget & Attack & $\varepsilon$ & mean $\|\delta\|_2$ & mean $\|\delta\|_\infty$ & mean $\|\delta\|_1$ \\
\midrule
\multirow{3}{*}{Extreme Low} & FGSM & 0.007843 & 0.1693 & 0.0078 & 3.6696 \\
 & $l_2$ & 0.125000 & 0.0965 & 0.0169 & 1.4967 \\
 & $l_\infty$ & 0.007843 & 0.0411 & 0.0071 & 0.6388 \\
\midrule
\multirow{3}{*}{Low} & FGSM & 0.015686 & 0.3348 & 0.0157 & 7.2238 \\
 & $l_2$ & 0.250000 & 0.1918 & 0.0339 & 2.9683 \\
 & $l_\infty$ & 0.015686 & 0.0820 & 0.0143 & 1.2742 \\
\midrule
\multirow{3}{*}{Medium} & FGSM & 0.031373 & 0.6655 & 0.0314 & 14.2656 \\
 & $l_2$ & 0.500000 & 0.3811 & 0.0677 & 5.8665 \\
 & $l_\infty$ & 0.031373 & 0.1631 & 0.0285 & 2.5302 \\
\midrule
\multirow{3}{*}{High} & FGSM & 0.062745 & 1.3250 & 0.0627 & 28.2429 \\
 & $l_2$ & 1.000000 & 0.7579 & 0.1349 & 11.5892 \\
 & $l_\infty$ & 0.062745 & 0.3241 & 0.0570 & 5.0035 \\
\midrule
\multirow{3}{*}{Extreme High} & FGSM & 0.125490 & 2.6336 & 0.1255 & 55.8307 \\
 & $l_2$ & 2.000000 & 1.5048 & 0.2690 & 22.8762 \\
 & $l_\infty$ & 0.125490 & 0.6445 & 0.1138 & 9.8859 \\
\bottomrule
\end{tabular}
\end{table}

\section{Complete results tables}
\label{app:complete_results}

Table \ref{tab:acc-full} shows the complete cross-attack accuracy measurements of all networks across all training regimes.
Tables \ref{tab:geom-raw} and \ref{tab:geom-phate} give the corresponding geometry in the raw $h_2$ space and in the 3-component M-PHATE embedding, each reporting the mean displacement, the clean--clean median used as the denominator, the resulting normalized displacement, $MMD^2$ and corresponding p-values.

\begin{table}[h]
\centering
\caption{Complete cross-attack accuracy grid. Every network is evaluated on the clean batch and under all three attacks at that budget; attacks are generated white-box against the network being evaluated. Mean $\pm$ s.d. over seeds 0, 1, 2.}
\label{tab:acc-full}
\small
\resizebox{\textwidth}{!}{%
\begin{tabular}{llccccccc}
\toprule
Budget & Trained & Clean & \multicolumn{3}{c}{Accuracy under attack} & \multicolumn{3}{c}{Gap (clean $-$ attacked)} \\
\cmidrule(lr){4-6} \cmidrule(lr){7-9}
 & & & FGSM & $l_2$ & $l_\infty$ & FGSM & $l_2$ & $l_\infty$ \\
\midrule
\multirow{4}{*}{Extreme Low} & Vanilla & 0.893 $\pm$ 0.002 & 0.855 $\pm$ 0.002 & 0.862 $\pm$ 0.007 & 0.882 $\pm$ 0.005 & 0.039 $\pm$ 0.002 & 0.031 $\pm$ 0.006 & 0.011 $\pm$ 0.003 \\
 & FGSM & 0.900 $\pm$ 0.003 & 0.869 $\pm$ 0.007 & 0.871 $\pm$ 0.006 & 0.890 $\pm$ 0.004 & 0.031 $\pm$ 0.004 & 0.030 $\pm$ 0.003 & 0.010 $\pm$ 0.002 \\
 & $l_2$ & 0.892 $\pm$ 0.005 & 0.863 $\pm$ 0.003 & 0.867 $\pm$ 0.003 & 0.882 $\pm$ 0.002 & 0.030 $\pm$ 0.005 & 0.025 $\pm$ 0.004 & 0.010 $\pm$ 0.004 \\
 & $l_\infty$ & 0.895 $\pm$ 0.007 & 0.862 $\pm$ 0.001 & 0.868 $\pm$ 0.004 & 0.884 $\pm$ 0.007 & 0.033 $\pm$ 0.008 & 0.027 $\pm$ 0.004 & 0.011 $\pm$ 0.002 \\
\midrule
\multirow{4}{*}{Low} & Vanilla & 0.893 $\pm$ 0.002 & 0.806 $\pm$ 0.002 & 0.828 $\pm$ 0.004 & 0.868 $\pm$ 0.008 & 0.087 $\pm$ 0.000 & 0.066 $\pm$ 0.005 & 0.025 $\pm$ 0.008 \\
 & FGSM & 0.907 $\pm$ 0.004 & 0.851 $\pm$ 0.005 & 0.854 $\pm$ 0.005 & 0.890 $\pm$ 0.004 & 0.056 $\pm$ 0.002 & 0.053 $\pm$ 0.003 & 0.018 $\pm$ 0.001 \\
 & $l_2$ & 0.898 $\pm$ 0.010 & 0.843 $\pm$ 0.008 & 0.852 $\pm$ 0.007 & 0.881 $\pm$ 0.011 & 0.055 $\pm$ 0.009 & 0.046 $\pm$ 0.006 & 0.017 $\pm$ 0.003 \\
 & $l_\infty$ & 0.895 $\pm$ 0.007 & 0.830 $\pm$ 0.005 & 0.841 $\pm$ 0.004 & 0.874 $\pm$ 0.001 & 0.065 $\pm$ 0.003 & 0.054 $\pm$ 0.003 & 0.021 $\pm$ 0.006 \\
\midrule
\multirow{4}{*}{Medium} & Vanilla & 0.893 $\pm$ 0.002 & 0.661 $\pm$ 0.019 & 0.725 $\pm$ 0.015 & 0.838 $\pm$ 0.004 & 0.232 $\pm$ 0.020 & 0.168 $\pm$ 0.016 & 0.055 $\pm$ 0.005 \\
 & FGSM & 0.909 $\pm$ 0.004 & 0.804 $\pm$ 0.011 & 0.802 $\pm$ 0.008 & 0.878 $\pm$ 0.007 & 0.105 $\pm$ 0.010 & 0.107 $\pm$ 0.007 & 0.030 $\pm$ 0.006 \\
 & $l_2$ & 0.908 $\pm$ 0.008 & 0.785 $\pm$ 0.001 & 0.796 $\pm$ 0.004 & 0.873 $\pm$ 0.005 & 0.123 $\pm$ 0.009 & 0.112 $\pm$ 0.008 & 0.035 $\pm$ 0.003 \\
 & $l_\infty$ & 0.899 $\pm$ 0.004 & 0.746 $\pm$ 0.012 & 0.771 $\pm$ 0.008 & 0.855 $\pm$ 0.005 & 0.153 $\pm$ 0.013 & 0.129 $\pm$ 0.010 & 0.044 $\pm$ 0.006 \\
\midrule
\multirow{4}{*}{High} & Vanilla & 0.893 $\pm$ 0.002 & 0.296 $\pm$ 0.030 & 0.430 $\pm$ 0.038 & 0.766 $\pm$ 0.005 & 0.597 $\pm$ 0.031 & 0.464 $\pm$ 0.039 & 0.127 $\pm$ 0.006 \\
 & FGSM & 0.907 $\pm$ 0.003 & 0.721 $\pm$ 0.014 & 0.668 $\pm$ 0.022 & 0.866 $\pm$ 0.006 & 0.186 $\pm$ 0.012 & 0.239 $\pm$ 0.021 & 0.041 $\pm$ 0.009 \\
 & $l_2$ & 0.916 $\pm$ 0.007 & 0.684 $\pm$ 0.004 & 0.696 $\pm$ 0.008 & 0.865 $\pm$ 0.010 & 0.231 $\pm$ 0.005 & 0.220 $\pm$ 0.001 & 0.050 $\pm$ 0.004 \\
 & $l_\infty$ & 0.909 $\pm$ 0.002 & 0.531 $\pm$ 0.017 & 0.589 $\pm$ 0.020 & 0.821 $\pm$ 0.006 & 0.378 $\pm$ 0.017 & 0.320 $\pm$ 0.020 & 0.089 $\pm$ 0.007 \\
\midrule
\multirow{4}{*}{Extreme High} & Vanilla & 0.893 $\pm$ 0.002 & 0.038 $\pm$ 0.002 & 0.092 $\pm$ 0.006 & 0.514 $\pm$ 0.034 & 0.856 $\pm$ 0.001 & 0.801 $\pm$ 0.007 & 0.380 $\pm$ 0.035 \\
 & FGSM & 0.859 $\pm$ 0.008 & 0.533 $\pm$ 0.014 & 0.389 $\pm$ 0.033 & 0.756 $\pm$ 0.019 & 0.326 $\pm$ 0.006 & 0.470 $\pm$ 0.025 & 0.102 $\pm$ 0.012 \\
 & $l_2$ & 0.897 $\pm$ 0.004 & 0.485 $\pm$ 0.015 & 0.499 $\pm$ 0.019 & 0.758 $\pm$ 0.003 & 0.412 $\pm$ 0.019 & 0.398 $\pm$ 0.022 & 0.139 $\pm$ 0.004 \\
 & $l_\infty$ & 0.914 $\pm$ 0.004 & 0.192 $\pm$ 0.023 & 0.248 $\pm$ 0.019 & 0.741 $\pm$ 0.004 & 0.721 $\pm$ 0.021 & 0.666 $\pm$ 0.015 & 0.173 $\pm$ 0.002 \\
\bottomrule
\end{tabular}}
\end{table}

\begin{table}[h]
\centering
\caption{Geometry in the raw 64-dimensional penultimate (h2) space. Mean displacement and median clean--clean pairwise distance are reported so normalized values can be recomputed as their ratio. Mean $\pm$ s.d. over seeds 0, 1, 2.}
\label{tab:geom-raw}
\tiny
\setlength{\tabcolsep}{1.5pt}
\begin{tabular}{llcccccc}
\toprule
Trained & Attack & Mean Disp. & \shortstack{Median\\Clean--Clean} & \shortstack{Norm.\\Disp.} & \shortstack{kNN\\Rec.} & MMD$^2$ & $p$ \\
\midrule
\multicolumn{8}{c}{\textbf{Extreme Low Budget}} \\
\midrule
Vanilla & FGSM & 1.144 $\pm$ 0.041 & 30.977 $\pm$ 0.787 & 0.0370 $\pm$ 0.0021 & 1.000 $\pm$ 0.000 & 0.00017 $\pm$ 0.00003 & 1.000 $\pm$ 0.000 \\
Vanilla & $l_2$ & 0.890 $\pm$ 0.038 & 30.977 $\pm$ 0.787 & 0.0288 $\pm$ 0.0018 & 1.000 $\pm$ 0.000 & 0.00007 $\pm$ 0.00002 & 1.000 $\pm$ 0.000 \\
Vanilla & $l_\infty$ & 0.376 $\pm$ 0.016 & 30.977 $\pm$ 0.787 & 0.0122 $\pm$ 0.0008 & 1.000 $\pm$ 0.000 & 0.00001 $\pm$ 0.00000 & 1.000 $\pm$ 0.000 \\
FGSM & FGSM & 0.871 $\pm$ 0.012 & 30.196 $\pm$ 0.449 & 0.0288 $\pm$ 0.0001 & 1.000 $\pm$ 0.000 & 0.00011 $\pm$ 0.00001 & 1.000 $\pm$ 0.000 \\
FGSM & $l_2$ & 0.751 $\pm$ 0.009 & 30.196 $\pm$ 0.449 & 0.0249 $\pm$ 0.0001 & 1.000 $\pm$ 0.000 & 0.00005 $\pm$ 0.00001 & 1.000 $\pm$ 0.000 \\
FGSM & $l_\infty$ & 0.289 $\pm$ 0.006 & 30.196 $\pm$ 0.449 & 0.0096 $\pm$ 0.0001 & 1.000 $\pm$ 0.000 & 0.00001 $\pm$ 0.00000 & 1.000 $\pm$ 0.000 \\
$l_2$ & FGSM & 0.949 $\pm$ 0.034 & 30.568 $\pm$ 0.322 & 0.0310 $\pm$ 0.0008 & 1.000 $\pm$ 0.000 & 0.00013 $\pm$ 0.00002 & 1.000 $\pm$ 0.000 \\
$l_2$ & $l_2$ & 0.782 $\pm$ 0.024 & 30.568 $\pm$ 0.322 & 0.0256 $\pm$ 0.0005 & 1.000 $\pm$ 0.000 & 0.00006 $\pm$ 0.00001 & 1.000 $\pm$ 0.000 \\
$l_2$ & $l_\infty$ & 0.328 $\pm$ 0.011 & 30.568 $\pm$ 0.322 & 0.0107 $\pm$ 0.0003 & 1.000 $\pm$ 0.000 & 0.00001 $\pm$ 0.00000 & 1.000 $\pm$ 0.000 \\
$l_\infty$ & FGSM & 0.961 $\pm$ 0.011 & 30.135 $\pm$ 0.340 & 0.0319 $\pm$ 0.0000 & 1.000 $\pm$ 0.000 & 0.00012 $\pm$ 0.00000 & 1.000 $\pm$ 0.000 \\
$l_\infty$ & $l_2$ & 0.783 $\pm$ 0.008 & 30.135 $\pm$ 0.340 & 0.0260 $\pm$ 0.0002 & 1.000 $\pm$ 0.000 & 0.00006 $\pm$ 0.00000 & 1.000 $\pm$ 0.000 \\
$l_\infty$ & $l_\infty$ & 0.326 $\pm$ 0.004 & 30.135 $\pm$ 0.340 & 0.0108 $\pm$ 0.0002 & 1.000 $\pm$ 0.000 & 0.00001 $\pm$ 0.00000 & 1.000 $\pm$ 0.000 \\
\midrule
\multicolumn{8}{c}{\textbf{Low Budget}} \\
\midrule
Vanilla & FGSM & 2.262 $\pm$ 0.081 & 30.977 $\pm$ 0.787 & 0.0731 $\pm$ 0.0041 & 1.000 $\pm$ 0.000 & 0.00068 $\pm$ 0.00010 & 0.918 $\pm$ 0.065 \\
Vanilla & $l_2$ & 1.768 $\pm$ 0.074 & 30.977 $\pm$ 0.787 & 0.0571 $\pm$ 0.0036 & 1.000 $\pm$ 0.000 & 0.00029 $\pm$ 0.00006 & 1.000 $\pm$ 0.001 \\
Vanilla & $l_\infty$ & 0.751 $\pm$ 0.033 & 30.977 $\pm$ 0.787 & 0.0243 $\pm$ 0.0015 & 1.000 $\pm$ 0.000 & 0.00005 $\pm$ 0.00001 & 1.000 $\pm$ 0.000 \\
FGSM & FGSM & 1.545 $\pm$ 0.043 & 30.158 $\pm$ 0.924 & 0.0512 $\pm$ 0.0002 & 1.000 $\pm$ 0.000 & 0.00037 $\pm$ 0.00004 & 0.999 $\pm$ 0.001 \\
FGSM & $l_2$ & 1.405 $\pm$ 0.039 & 30.158 $\pm$ 0.924 & 0.0466 $\pm$ 0.0002 & 1.000 $\pm$ 0.000 & 0.00018 $\pm$ 0.00002 & 1.000 $\pm$ 0.000 \\
FGSM & $l_\infty$ & 0.528 $\pm$ 0.016 & 30.158 $\pm$ 0.924 & 0.0175 $\pm$ 0.0003 & 1.000 $\pm$ 0.000 & 0.00003 $\pm$ 0.00000 & 1.000 $\pm$ 0.000 \\
$l_2$ & FGSM & 1.716 $\pm$ 0.033 & 30.483 $\pm$ 0.424 & 0.0563 $\pm$ 0.0013 & 1.000 $\pm$ 0.000 & 0.00044 $\pm$ 0.00007 & 0.996 $\pm$ 0.003 \\
$l_2$ & $l_2$ & 1.454 $\pm$ 0.022 & 30.483 $\pm$ 0.424 & 0.0477 $\pm$ 0.0007 & 1.000 $\pm$ 0.000 & 0.00019 $\pm$ 0.00003 & 1.000 $\pm$ 0.000 \\
$l_2$ & $l_\infty$ & 0.602 $\pm$ 0.009 & 30.483 $\pm$ 0.424 & 0.0197 $\pm$ 0.0002 & 1.000 $\pm$ 0.000 & 0.00003 $\pm$ 0.00000 & 1.000 $\pm$ 0.000 \\
$l_\infty$ & FGSM & 1.792 $\pm$ 0.048 & 30.046 $\pm$ 0.504 & 0.0596 $\pm$ 0.0006 & 1.000 $\pm$ 0.000 & 0.00046 $\pm$ 0.00003 & 0.994 $\pm$ 0.004 \\
$l_\infty$ & $l_2$ & 1.482 $\pm$ 0.036 & 30.046 $\pm$ 0.504 & 0.0493 $\pm$ 0.0005 & 1.000 $\pm$ 0.000 & 0.00019 $\pm$ 0.00002 & 1.000 $\pm$ 0.000 \\
$l_\infty$ & $l_\infty$ & 0.619 $\pm$ 0.011 & 30.046 $\pm$ 0.504 & 0.0206 $\pm$ 0.0003 & 1.000 $\pm$ 0.000 & 0.00003 $\pm$ 0.00000 & 1.000 $\pm$ 0.000 \\
\midrule
\multicolumn{8}{c}{\textbf{Medium Budget}} \\
\midrule
Vanilla & FGSM & 4.499 $\pm$ 0.158 & 30.977 $\pm$ 0.787 & 0.1454 $\pm$ 0.0080 & 0.996 $\pm$ 0.003 & 0.00263 $\pm$ 0.00035 & 0.017 $\pm$ 0.012 \\
Vanilla & $l_2$ & 3.511 $\pm$ 0.147 & 30.977 $\pm$ 0.787 & 0.1135 $\pm$ 0.0071 & 0.999 $\pm$ 0.001 & 0.00112 $\pm$ 0.00022 & 0.526 $\pm$ 0.198 \\
Vanilla & $l_\infty$ & 1.493 $\pm$ 0.065 & 30.977 $\pm$ 0.787 & 0.0482 $\pm$ 0.0031 & 1.000 $\pm$ 0.000 & 0.00021 $\pm$ 0.00004 & 1.000 $\pm$ 0.000 \\
FGSM & FGSM & 2.570 $\pm$ 0.056 & 28.917 $\pm$ 0.520 & 0.0889 $\pm$ 0.0016 & 1.000 $\pm$ 0.000 & 0.00105 $\pm$ 0.00015 & 0.575 $\pm$ 0.140 \\
FGSM & $l_2$ & 2.519 $\pm$ 0.037 & 28.917 $\pm$ 0.520 & 0.0871 $\pm$ 0.0016 & 1.000 $\pm$ 0.000 & 0.00063 $\pm$ 0.00010 & 0.948 $\pm$ 0.053 \\
FGSM & $l_\infty$ & 0.885 $\pm$ 0.018 & 28.917 $\pm$ 0.520 & 0.0306 $\pm$ 0.0007 & 1.000 $\pm$ 0.000 & 0.00008 $\pm$ 0.00001 & 1.000 $\pm$ 0.000 \\
$l_2$ & FGSM & 2.876 $\pm$ 0.064 & 28.991 $\pm$ 0.314 & 0.0992 $\pm$ 0.0014 & 1.000 $\pm$ 0.000 & 0.00138 $\pm$ 0.00015 & 0.318 $\pm$ 0.111 \\
$l_2$ & $l_2$ & 2.515 $\pm$ 0.059 & 28.991 $\pm$ 0.314 & 0.0867 $\pm$ 0.0013 & 1.000 $\pm$ 0.000 & 0.00064 $\pm$ 0.00008 & 0.946 $\pm$ 0.040 \\
$l_2$ & $l_\infty$ & 1.018 $\pm$ 0.024 & 28.991 $\pm$ 0.314 & 0.0351 $\pm$ 0.0005 & 1.000 $\pm$ 0.000 & 0.00011 $\pm$ 0.00001 & 1.000 $\pm$ 0.000 \\
$l_\infty$ & FGSM & 3.377 $\pm$ 0.093 & 30.431 $\pm$ 0.536 & 0.1110 $\pm$ 0.0011 & 1.000 $\pm$ 0.000 & 0.00154 $\pm$ 0.00017 & 0.215 $\pm$ 0.090 \\
$l_\infty$ & $l_2$ & 2.839 $\pm$ 0.069 & 30.431 $\pm$ 0.536 & 0.0933 $\pm$ 0.0008 & 1.000 $\pm$ 0.000 & 0.00060 $\pm$ 0.00009 & 0.959 $\pm$ 0.031 \\
$l_\infty$ & $l_\infty$ & 1.187 $\pm$ 0.026 & 30.431 $\pm$ 0.536 & 0.0390 $\pm$ 0.0003 & 1.000 $\pm$ 0.000 & 0.00011 $\pm$ 0.00002 & 1.000 $\pm$ 0.000 \\
\midrule
\multicolumn{8}{c}{\textbf{High Budget}} \\
\midrule
Vanilla & FGSM & 8.967 $\pm$ 0.299 & 30.977 $\pm$ 0.787 & 0.2897 $\pm$ 0.0152 & 0.874 $\pm$ 0.063 & 0.00964 $\pm$ 0.00125 & 0.000 $\pm$ 0.000 \\
Vanilla & $l_2$ & 6.963 $\pm$ 0.285 & 30.977 $\pm$ 0.787 & 0.2250 $\pm$ 0.0139 & 0.949 $\pm$ 0.033 & 0.00410 $\pm$ 0.00082 & 0.000 $\pm$ 0.001 \\
Vanilla & $l_\infty$ & 2.965 $\pm$ 0.129 & 30.977 $\pm$ 0.787 & 0.0958 $\pm$ 0.0061 & 1.000 $\pm$ 0.001 & 0.00083 $\pm$ 0.00013 & 0.788 $\pm$ 0.140 \\
FGSM & FGSM & 3.357 $\pm$ 0.194 & 23.117 $\pm$ 1.095 & 0.1453 $\pm$ 0.0067 & 0.993 $\pm$ 0.013 & 0.00320 $\pm$ 0.00081 & 0.009 $\pm$ 0.014 \\
FGSM & $l_2$ & 3.632 $\pm$ 0.139 & 23.117 $\pm$ 1.095 & 0.1572 $\pm$ 0.0057 & 0.986 $\pm$ 0.016 & 0.00259 $\pm$ 0.00053 & 0.025 $\pm$ 0.026 \\
FGSM & $l_\infty$ & 1.130 $\pm$ 0.070 & 23.117 $\pm$ 1.095 & 0.0489 $\pm$ 0.0027 & 1.000 $\pm$ 0.000 & 0.00028 $\pm$ 0.00007 & 1.000 $\pm$ 0.000 \\
$l_2$ & FGSM & 3.959 $\pm$ 0.109 & 23.396 $\pm$ 0.141 & 0.1692 $\pm$ 0.0049 & 0.994 $\pm$ 0.007 & 0.00438 $\pm$ 0.00077 & 0.001 $\pm$ 0.001 \\
$l_2$ & $l_2$ & 3.618 $\pm$ 0.070 & 23.396 $\pm$ 0.141 & 0.1546 $\pm$ 0.0033 & 0.989 $\pm$ 0.013 & 0.00242 $\pm$ 0.00045 & 0.033 $\pm$ 0.041 \\
$l_2$ & $l_\infty$ & 1.438 $\pm$ 0.036 & 23.396 $\pm$ 0.141 & 0.0615 $\pm$ 0.0016 & 1.000 $\pm$ 0.000 & 0.00043 $\pm$ 0.00009 & 0.998 $\pm$ 0.003 \\
$l_\infty$ & FGSM & 6.088 $\pm$ 0.209 & 29.736 $\pm$ 0.814 & 0.2047 $\pm$ 0.0030 & 0.992 $\pm$ 0.003 & 0.00574 $\pm$ 0.00064 & 0.000 $\pm$ 0.000 \\
$l_\infty$ & $l_2$ & 5.210 $\pm$ 0.171 & 29.736 $\pm$ 0.814 & 0.1752 $\pm$ 0.0026 & 0.994 $\pm$ 0.001 & 0.00223 $\pm$ 0.00024 & 0.040 $\pm$ 0.037 \\
$l_\infty$ & $l_\infty$ & 2.137 $\pm$ 0.065 & 29.736 $\pm$ 0.814 & 0.0719 $\pm$ 0.0015 & 1.000 $\pm$ 0.000 & 0.00041 $\pm$ 0.00004 & 0.997 $\pm$ 0.003 \\
\midrule
\multicolumn{8}{c}{\textbf{Extreme High Budget}} \\
\midrule
Vanilla & FGSM & 17.525 $\pm$ 0.487 & 30.977 $\pm$ 0.787 & 0.5661 $\pm$ 0.0247 & 0.240 $\pm$ 0.030 & 0.03412 $\pm$ 0.00457 & 0.000 $\pm$ 0.000 \\
Vanilla & $l_2$ & 13.589 $\pm$ 0.501 & 30.977 $\pm$ 0.787 & 0.4390 $\pm$ 0.0250 & 0.508 $\pm$ 0.052 & 0.01521 $\pm$ 0.00349 & 0.000 $\pm$ 0.000 \\
Vanilla & $l_\infty$ & 5.885 $\pm$ 0.253 & 30.977 $\pm$ 0.787 & 0.1902 $\pm$ 0.0119 & 0.983 $\pm$ 0.010 & 0.00308 $\pm$ 0.00049 & 0.006 $\pm$ 0.006 \\
FGSM & FGSM & 3.152 $\pm$ 0.182 & 13.333 $\pm$ 0.688 & 0.2364 $\pm$ 0.0068 & 0.945 $\pm$ 0.032 & 0.00913 $\pm$ 0.00132 & 0.000 $\pm$ 0.000 \\
FGSM & $l_2$ & 3.887 $\pm$ 0.169 & 13.333 $\pm$ 0.688 & 0.2916 $\pm$ 0.0027 & 0.804 $\pm$ 0.022 & 0.00891 $\pm$ 0.00094 & 0.000 $\pm$ 0.000 \\
FGSM & $l_\infty$ & 1.108 $\pm$ 0.071 & 13.333 $\pm$ 0.688 & 0.0831 $\pm$ 0.0026 & 1.000 $\pm$ 0.000 & 0.00091 $\pm$ 0.00011 & 0.711 $\pm$ 0.111 \\
$l_2$ & FGSM & 4.077 $\pm$ 0.069 & 14.473 $\pm$ 0.350 & 0.2817 $\pm$ 0.0023 & 0.887 $\pm$ 0.014 & 0.01461 $\pm$ 0.00072 & 0.000 $\pm$ 0.000 \\
$l_2$ & $l_2$ & 3.719 $\pm$ 0.046 & 14.473 $\pm$ 0.350 & 0.2570 $\pm$ 0.0035 & 0.908 $\pm$ 0.003 & 0.00717 $\pm$ 0.00039 & 0.000 $\pm$ 0.000 \\
$l_2$ & $l_\infty$ & 1.576 $\pm$ 0.011 & 14.473 $\pm$ 0.350 & 0.1089 $\pm$ 0.0019 & 0.995 $\pm$ 0.006 & 0.00152 $\pm$ 0.00010 & 0.214 $\pm$ 0.055 \\
$l_\infty$ & FGSM & 9.052 $\pm$ 0.403 & 25.573 $\pm$ 1.730 & 0.3543 $\pm$ 0.0084 & 0.762 $\pm$ 0.036 & 0.01708 $\pm$ 0.00081 & 0.000 $\pm$ 0.000 \\
$l_\infty$ & $l_2$ & 7.970 $\pm$ 0.324 & 25.573 $\pm$ 1.730 & 0.3120 $\pm$ 0.0082 & 0.858 $\pm$ 0.038 & 0.00647 $\pm$ 0.00070 & 0.000 $\pm$ 0.000 \\
$l_\infty$ & $l_\infty$ & 3.203 $\pm$ 0.120 & 25.573 $\pm$ 1.730 & 0.1254 $\pm$ 0.0040 & 0.997 $\pm$ 0.004 & 0.00143 $\pm$ 0.00017 & 0.286 $\pm$ 0.117 \\
\bottomrule
\end{tabular}
\end{table}

\begin{table}[t]
\centering
\caption{Geometry in the 3-component M-PHATE embedding of the joint clean/perturbed tensor. Columns match Table~\ref{tab:geom-raw}. Mean $\pm$ s.d. over seeds 0, 1, 2.}
\label{tab:geom-phate}
\tiny
\setlength{\tabcolsep}{1.5pt}
\begin{tabular}{llcccccc}
\toprule
Trained & Attack & Mean Disp. & \shortstack{Median\\Clean--Clean} & \shortstack{Norm.\\Disp.} & \shortstack{kNN\\Rec.} & MMD$^2$ & $p$ \\
\midrule
\multicolumn{8}{c}{\textbf{Extreme Low Budget}} \\
\midrule
Vanilla & FGSM & 0.00169 $\pm$ 0.00028 & 0.02985 $\pm$ 0.00033 & 0.0565 $\pm$ 0.0088 & 0.899 $\pm$ 0.037 & 0.00125 $\pm$ 0.00076 & 0.432 $\pm$ 0.405 \\
Vanilla & $l_2$ & 0.00150 $\pm$ 0.00033 & 0.02980 $\pm$ 0.00031 & 0.0503 $\pm$ 0.0105 & 0.912 $\pm$ 0.039 & 0.00126 $\pm$ 0.00099 & 0.480 $\pm$ 0.448 \\
Vanilla & $l_\infty$ & 0.00097 $\pm$ 0.00027 & 0.02969 $\pm$ 0.00034 & 0.0328 $\pm$ 0.0087 & 0.921 $\pm$ 0.026 & 0.00087 $\pm$ 0.00059 & 0.588 $\pm$ 0.365 \\
FGSM & FGSM & 0.00113 $\pm$ 0.00015 & 0.02971 $\pm$ 0.00026 & 0.0381 $\pm$ 0.0050 & 0.953 $\pm$ 0.018 & 0.00040 $\pm$ 0.00023 & 0.891 $\pm$ 0.139 \\
FGSM & $l_2$ & 0.00119 $\pm$ 0.00020 & 0.02969 $\pm$ 0.00025 & 0.0400 $\pm$ 0.0067 & 0.936 $\pm$ 0.014 & 0.00059 $\pm$ 0.00042 & 0.755 $\pm$ 0.305 \\
FGSM & $l_\infty$ & 0.00076 $\pm$ 0.00023 & 0.02950 $\pm$ 0.00017 & 0.0259 $\pm$ 0.0077 & 0.936 $\pm$ 0.029 & 0.00057 $\pm$ 0.00042 & 0.736 $\pm$ 0.232 \\
$l_2$ & FGSM & 0.00129 $\pm$ 0.00023 & 0.02981 $\pm$ 0.00084 & 0.0434 $\pm$ 0.0068 & 0.945 $\pm$ 0.028 & 0.00066 $\pm$ 0.00041 & 0.723 $\pm$ 0.270 \\
$l_2$ & $l_2$ & 0.00116 $\pm$ 0.00021 & 0.02977 $\pm$ 0.00080 & 0.0389 $\pm$ 0.0062 & 0.946 $\pm$ 0.027 & 0.00053 $\pm$ 0.00042 & 0.804 $\pm$ 0.278 \\
$l_2$ & $l_\infty$ & 0.00080 $\pm$ 0.00027 & 0.02965 $\pm$ 0.00079 & 0.0271 $\pm$ 0.0090 & 0.939 $\pm$ 0.027 & 0.00063 $\pm$ 0.00047 & 0.712 $\pm$ 0.249 \\
$l_\infty$ & FGSM & 0.00128 $\pm$ 0.00011 & 0.03018 $\pm$ 0.00011 & 0.0425 $\pm$ 0.0036 & 0.939 $\pm$ 0.020 & 0.00047 $\pm$ 0.00014 & 0.852 $\pm$ 0.105 \\
$l_\infty$ & $l_2$ & 0.00117 $\pm$ 0.00017 & 0.03015 $\pm$ 0.00014 & 0.0388 $\pm$ 0.0054 & 0.941 $\pm$ 0.031 & 0.00049 $\pm$ 0.00027 & 0.809 $\pm$ 0.188 \\
$l_\infty$ & $l_\infty$ & 0.00056 $\pm$ 0.00009 & 0.03000 $\pm$ 0.00019 & 0.0187 $\pm$ 0.0029 & 0.970 $\pm$ 0.024 & 0.00013 $\pm$ 0.00007 & 0.997 $\pm$ 0.004 \\
\midrule
\multicolumn{8}{c}{\textbf{Low Budget}} \\
\midrule
Vanilla & FGSM & 0.00291 $\pm$ 0.00033 & 0.03005 $\pm$ 0.00046 & 0.0966 $\pm$ 0.0096 & 0.783 $\pm$ 0.015 & 0.00270 $\pm$ 0.00135 & 0.123 $\pm$ 0.178 \\
Vanilla & $l_2$ & 0.00256 $\pm$ 0.00033 & 0.02997 $\pm$ 0.00037 & 0.0853 $\pm$ 0.0100 & 0.817 $\pm$ 0.029 & 0.00262 $\pm$ 0.00140 & 0.148 $\pm$ 0.213 \\
Vanilla & $l_\infty$ & 0.00141 $\pm$ 0.00030 & 0.02980 $\pm$ 0.00032 & 0.0472 $\pm$ 0.0096 & 0.913 $\pm$ 0.031 & 0.00126 $\pm$ 0.00084 & 0.438 $\pm$ 0.461 \\
FGSM & FGSM & 0.00192 $\pm$ 0.00019 & 0.03000 $\pm$ 0.00047 & 0.0640 $\pm$ 0.0056 & 0.913 $\pm$ 0.030 & 0.00096 $\pm$ 0.00041 & 0.531 $\pm$ 0.278 \\
FGSM & $l_2$ & 0.00183 $\pm$ 0.00021 & 0.02997 $\pm$ 0.00048 & 0.0609 $\pm$ 0.0060 & 0.916 $\pm$ 0.036 & 0.00088 $\pm$ 0.00042 & 0.573 $\pm$ 0.300 \\
FGSM & $l_\infty$ & 0.00081 $\pm$ 0.00014 & 0.02971 $\pm$ 0.00046 & 0.0271 $\pm$ 0.0044 & 0.971 $\pm$ 0.035 & 0.00025 $\pm$ 0.00017 & 0.963 $\pm$ 0.055 \\
$l_2$ & FGSM & 0.00217 $\pm$ 0.00027 & 0.03002 $\pm$ 0.00083 & 0.0723 $\pm$ 0.0073 & 0.875 $\pm$ 0.032 & 0.00145 $\pm$ 0.00048 & 0.283 $\pm$ 0.154 \\
$l_2$ & $l_2$ & 0.00192 $\pm$ 0.00018 & 0.02995 $\pm$ 0.00081 & 0.0640 $\pm$ 0.0045 & 0.902 $\pm$ 0.030 & 0.00105 $\pm$ 0.00016 & 0.477 $\pm$ 0.099 \\
$l_2$ & $l_\infty$ & 0.00098 $\pm$ 0.00009 & 0.02970 $\pm$ 0.00076 & 0.0328 $\pm$ 0.0022 & 0.952 $\pm$ 0.017 & 0.00037 $\pm$ 0.00006 & 0.929 $\pm$ 0.032 \\
$l_\infty$ & FGSM & 0.00218 $\pm$ 0.00003 & 0.03049 $\pm$ 0.00004 & 0.0716 $\pm$ 0.0009 & 0.875 $\pm$ 0.008 & 0.00113 $\pm$ 0.00010 & 0.408 $\pm$ 0.062 \\
$l_\infty$ & $l_2$ & 0.00194 $\pm$ 0.00013 & 0.03041 $\pm$ 0.00006 & 0.0637 $\pm$ 0.0040 & 0.900 $\pm$ 0.016 & 0.00088 $\pm$ 0.00024 & 0.542 $\pm$ 0.165 \\
$l_\infty$ & $l_\infty$ & 0.00088 $\pm$ 0.00012 & 0.03010 $\pm$ 0.00004 & 0.0292 $\pm$ 0.0040 & 0.976 $\pm$ 0.034 & 0.00023 $\pm$ 0.00011 & 0.974 $\pm$ 0.035 \\
\midrule
\multicolumn{8}{c}{\textbf{Medium Budget}} \\
\midrule
Vanilla & FGSM & 0.00492 $\pm$ 0.00039 & 0.03046 $\pm$ 0.00065 & 0.1613 $\pm$ 0.0096 & 0.523 $\pm$ 0.005 & 0.00606 $\pm$ 0.00242 & 0.005 $\pm$ 0.008 \\
Vanilla & $l_2$ & 0.00423 $\pm$ 0.00043 & 0.03023 $\pm$ 0.00052 & 0.1397 $\pm$ 0.0117 & 0.615 $\pm$ 0.013 & 0.00507 $\pm$ 0.00231 & 0.010 $\pm$ 0.017 \\
Vanilla & $l_\infty$ & 0.00236 $\pm$ 0.00029 & 0.02998 $\pm$ 0.00037 & 0.0786 $\pm$ 0.0087 & 0.855 $\pm$ 0.028 & 0.00259 $\pm$ 0.00124 & 0.132 $\pm$ 0.194 \\
FGSM & FGSM & 0.00307 $\pm$ 0.00012 & 0.03028 $\pm$ 0.00032 & 0.1015 $\pm$ 0.0050 & 0.795 $\pm$ 0.027 & 0.00202 $\pm$ 0.00013 & 0.112 $\pm$ 0.018 \\
FGSM & $l_2$ & 0.00303 $\pm$ 0.00016 & 0.03022 $\pm$ 0.00027 & 0.1001 $\pm$ 0.0059 & 0.779 $\pm$ 0.031 & 0.00180 $\pm$ 0.00028 & 0.165 $\pm$ 0.063 \\
FGSM & $l_\infty$ & 0.00125 $\pm$ 0.00011 & 0.02988 $\pm$ 0.00027 & 0.0418 $\pm$ 0.0040 & 0.959 $\pm$ 0.040 & 0.00045 $\pm$ 0.00012 & 0.876 $\pm$ 0.083 \\
$l_2$ & FGSM & 0.00364 $\pm$ 0.00038 & 0.03062 $\pm$ 0.00029 & 0.1189 $\pm$ 0.0129 & 0.705 $\pm$ 0.026 & 0.00368 $\pm$ 0.00141 & 0.020 $\pm$ 0.020 \\
$l_2$ & $l_2$ & 0.00320 $\pm$ 0.00036 & 0.03039 $\pm$ 0.00032 & 0.1053 $\pm$ 0.0124 & 0.751 $\pm$ 0.034 & 0.00240 $\pm$ 0.00102 & 0.121 $\pm$ 0.111 \\
$l_2$ & $l_\infty$ & 0.00159 $\pm$ 0.00035 & 0.03001 $\pm$ 0.00021 & 0.0529 $\pm$ 0.0120 & 0.942 $\pm$ 0.036 & 0.00096 $\pm$ 0.00078 & 0.592 $\pm$ 0.413 \\
$l_\infty$ & FGSM & 0.00380 $\pm$ 0.00008 & 0.03086 $\pm$ 0.00020 & 0.1231 $\pm$ 0.0023 & 0.671 $\pm$ 0.003 & 0.00315 $\pm$ 0.00008 & 0.018 $\pm$ 0.002 \\
$l_\infty$ & $l_2$ & 0.00338 $\pm$ 0.00018 & 0.03070 $\pm$ 0.00023 & 0.1101 $\pm$ 0.0053 & 0.720 $\pm$ 0.020 & 0.00233 $\pm$ 0.00046 & 0.072 $\pm$ 0.053 \\
$l_\infty$ & $l_\infty$ & 0.00161 $\pm$ 0.00014 & 0.03025 $\pm$ 0.00024 & 0.0532 $\pm$ 0.0045 & 0.927 $\pm$ 0.041 & 0.00069 $\pm$ 0.00020 & 0.703 $\pm$ 0.156 \\
\midrule
\multicolumn{8}{c}{\textbf{High Budget}} \\
\midrule
Vanilla & FGSM & 0.00816 $\pm$ 0.00026 & 0.03087 $\pm$ 0.00048 & 0.2644 $\pm$ 0.0047 & 0.247 $\pm$ 0.025 & 0.01396 $\pm$ 0.00250 & 0.000 $\pm$ 0.000 \\
Vanilla & $l_2$ & 0.00692 $\pm$ 0.00037 & 0.03035 $\pm$ 0.00061 & 0.2278 $\pm$ 0.0081 & 0.348 $\pm$ 0.020 & 0.01156 $\pm$ 0.00179 & 0.000 $\pm$ 0.000 \\
Vanilla & $l_\infty$ & 0.00382 $\pm$ 0.00033 & 0.03022 $\pm$ 0.00055 & 0.1262 $\pm$ 0.0086 & 0.665 $\pm$ 0.019 & 0.00461 $\pm$ 0.00182 & 0.014 $\pm$ 0.023 \\
FGSM & FGSM & 0.00436 $\pm$ 0.00005 & 0.03040 $\pm$ 0.00031 & 0.1436 $\pm$ 0.0030 & 0.562 $\pm$ 0.019 & 0.00400 $\pm$ 0.00023 & 0.008 $\pm$ 0.003 \\
FGSM & $l_2$ & 0.00441 $\pm$ 0.00017 & 0.03040 $\pm$ 0.00027 & 0.1451 $\pm$ 0.0069 & 0.540 $\pm$ 0.038 & 0.00351 $\pm$ 0.00087 & 0.015 $\pm$ 0.011 \\
FGSM & $l_\infty$ & 0.00173 $\pm$ 0.00011 & 0.02976 $\pm$ 0.00020 & 0.0582 $\pm$ 0.0041 & 0.948 $\pm$ 0.015 & 0.00082 $\pm$ 0.00023 & 0.601 $\pm$ 0.164 \\
$l_2$ & FGSM & 0.00505 $\pm$ 0.00042 & 0.03126 $\pm$ 0.00048 & 0.1615 $\pm$ 0.0119 & 0.477 $\pm$ 0.063 & 0.00503 $\pm$ 0.00143 & 0.003 $\pm$ 0.004 \\
$l_2$ & $l_2$ & 0.00458 $\pm$ 0.00037 & 0.03102 $\pm$ 0.00052 & 0.1477 $\pm$ 0.0103 & 0.504 $\pm$ 0.066 & 0.00381 $\pm$ 0.00158 & 0.035 $\pm$ 0.057 \\
$l_2$ & $l_\infty$ & 0.00221 $\pm$ 0.00017 & 0.03036 $\pm$ 0.00051 & 0.0727 $\pm$ 0.0042 & 0.883 $\pm$ 0.034 & 0.00138 $\pm$ 0.00039 & 0.312 $\pm$ 0.149 \\
$l_\infty$ & FGSM & 0.00601 $\pm$ 0.00016 & 0.03102 $\pm$ 0.00042 & 0.1937 $\pm$ 0.0034 & 0.383 $\pm$ 0.042 & 0.00619 $\pm$ 0.00107 & 0.000 $\pm$ 0.001 \\
$l_\infty$ & $l_2$ & 0.00536 $\pm$ 0.00011 & 0.03069 $\pm$ 0.00037 & 0.1746 $\pm$ 0.0014 & 0.436 $\pm$ 0.039 & 0.00420 $\pm$ 0.00041 & 0.003 $\pm$ 0.002 \\
$l_\infty$ & $l_\infty$ & 0.00272 $\pm$ 0.00008 & 0.03042 $\pm$ 0.00030 & 0.0894 $\pm$ 0.0026 & 0.837 $\pm$ 0.019 & 0.00160 $\pm$ 0.00024 & 0.218 $\pm$ 0.080 \\
\midrule
\multicolumn{8}{c}{\textbf{Extreme High Budget}} \\
\midrule
Vanilla & FGSM & 0.01251 $\pm$ 0.00033 & 0.03022 $\pm$ 0.00044 & 0.4140 $\pm$ 0.0144 & 0.108 $\pm$ 0.012 & 0.02170 $\pm$ 0.00072 & 0.000 $\pm$ 0.000 \\
Vanilla & $l_2$ & 0.01082 $\pm$ 0.00014 & 0.03021 $\pm$ 0.00050 & 0.3583 $\pm$ 0.0080 & 0.160 $\pm$ 0.025 & 0.02063 $\pm$ 0.00063 & 0.000 $\pm$ 0.000 \\
Vanilla & $l_\infty$ & 0.00616 $\pm$ 0.00032 & 0.03042 $\pm$ 0.00054 & 0.2024 $\pm$ 0.0070 & 0.399 $\pm$ 0.011 & 0.00901 $\pm$ 0.00233 & 0.000 $\pm$ 0.000 \\
FGSM & FGSM & 0.00613 $\pm$ 0.00009 & 0.02984 $\pm$ 0.00061 & 0.2056 $\pm$ 0.0034 & 0.371 $\pm$ 0.016 & 0.00576 $\pm$ 0.00055 & 0.001 $\pm$ 0.001 \\
FGSM & $l_2$ & 0.00745 $\pm$ 0.00025 & 0.03042 $\pm$ 0.00057 & 0.2450 $\pm$ 0.0038 & 0.291 $\pm$ 0.031 & 0.00758 $\pm$ 0.00177 & 0.000 $\pm$ 0.000 \\
FGSM & $l_\infty$ & 0.00248 $\pm$ 0.00006 & 0.02978 $\pm$ 0.00046 & 0.0832 $\pm$ 0.0023 & 0.860 $\pm$ 0.018 & 0.00136 $\pm$ 0.00014 & 0.284 $\pm$ 0.061 \\
$l_2$ & FGSM & 0.00753 $\pm$ 0.00019 & 0.03220 $\pm$ 0.00039 & 0.2338 $\pm$ 0.0075 & 0.225 $\pm$ 0.013 & 0.01125 $\pm$ 0.00254 & 0.000 $\pm$ 0.000 \\
$l_2$ & $l_2$ & 0.00672 $\pm$ 0.00021 & 0.03197 $\pm$ 0.00032 & 0.2102 $\pm$ 0.0085 & 0.252 $\pm$ 0.010 & 0.00758 $\pm$ 0.00198 & 0.000 $\pm$ 0.001 \\
$l_2$ & $l_\infty$ & 0.00325 $\pm$ 0.00030 & 0.03110 $\pm$ 0.00024 & 0.1045 $\pm$ 0.0091 & 0.658 $\pm$ 0.021 & 0.00236 $\pm$ 0.00045 & 0.078 $\pm$ 0.055 \\
$l_\infty$ & FGSM & 0.00952 $\pm$ 0.00012 & 0.03171 $\pm$ 0.00042 & 0.3004 $\pm$ 0.0042 & 0.138 $\pm$ 0.025 & 0.02139 $\pm$ 0.00051 & 0.000 $\pm$ 0.000 \\
$l_\infty$ & $l_2$ & 0.00833 $\pm$ 0.00002 & 0.03145 $\pm$ 0.00017 & 0.2648 $\pm$ 0.0019 & 0.164 $\pm$ 0.021 & 0.01402 $\pm$ 0.00193 & 0.000 $\pm$ 0.000 \\
$l_\infty$ & $l_\infty$ & 0.00410 $\pm$ 0.00038 & 0.03075 $\pm$ 0.00025 & 0.1333 $\pm$ 0.0112 & 0.597 $\pm$ 0.039 & 0.00363 $\pm$ 0.00160 & 0.030 $\pm$ 0.039 \\
\bottomrule
\end{tabular}
\end{table}

\section{Additional figures}
\label{app:additional_figures}

\subsection{M-PHATE embeddings across training regimes and adversarial budgets }
\label{app:mphate-embeddings}

\begin{figure}
  \centering
  \caption{M-PHATE embeddings of training trajectories for layers $h_1$ (top block) and $h_2$ (bottom block), with one row per network and one column per perturbation budget. Each point is one clean test image at one epoch, colored by epoch. The vanilla row is identical across all five columns, since that network never sees an attack.}
    \label{fig:all_mphate_traj}
  \includegraphics[width=1.0\linewidth]{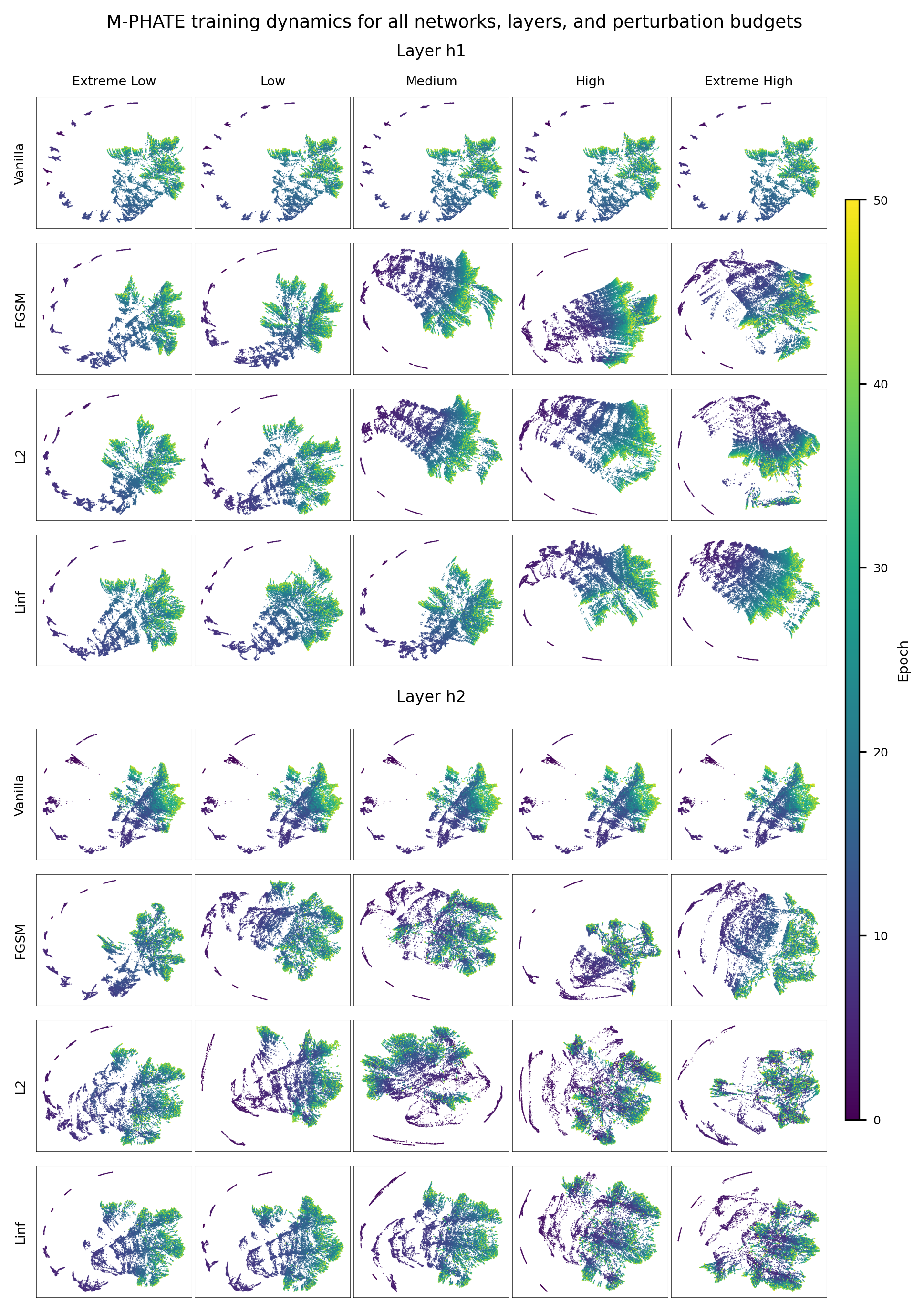}
\end{figure}

Figure \ref{fig:all_mphate_traj} shows the M-PHATE trajectories of all networks across 50 epochs. 
Clean test samples are fed through at every epoch. 
Throughout these additional examples, the pattern noted in the main text still holds: the first hidden layers ($h_1$) of each of these networks appears to have more stable and consistent trajectory evolution throughout training than the second hidden layers ($h_2$), whose trajectory is hard to follow from epoch to epoch. 
For the adversarially trained networks, this becomes more pronounced at larger perturbation budgets, where data points across epochs do not progress in a similar way as $h_1$.
The vanilla row is identical across all budgets.

\subsection{Joint M-PHATE embeddings across training regimes and adversarial budgets }
\label{app:phate-embeddings}

\begin{figure}[h]
  \centering
  \caption{Embeddings of the vanilla network's penultimate layer. Joint M-PHATE embeddings of clean (circles) and perturbed (crosses)
penultimate-layer representations of 1000 MNIST test images, colored by digit class,
across attacks (rows) and perturbation budgets (columns).}%
  \label{fig:vanilla_network_joint}
  \includegraphics[width=1.0\linewidth]{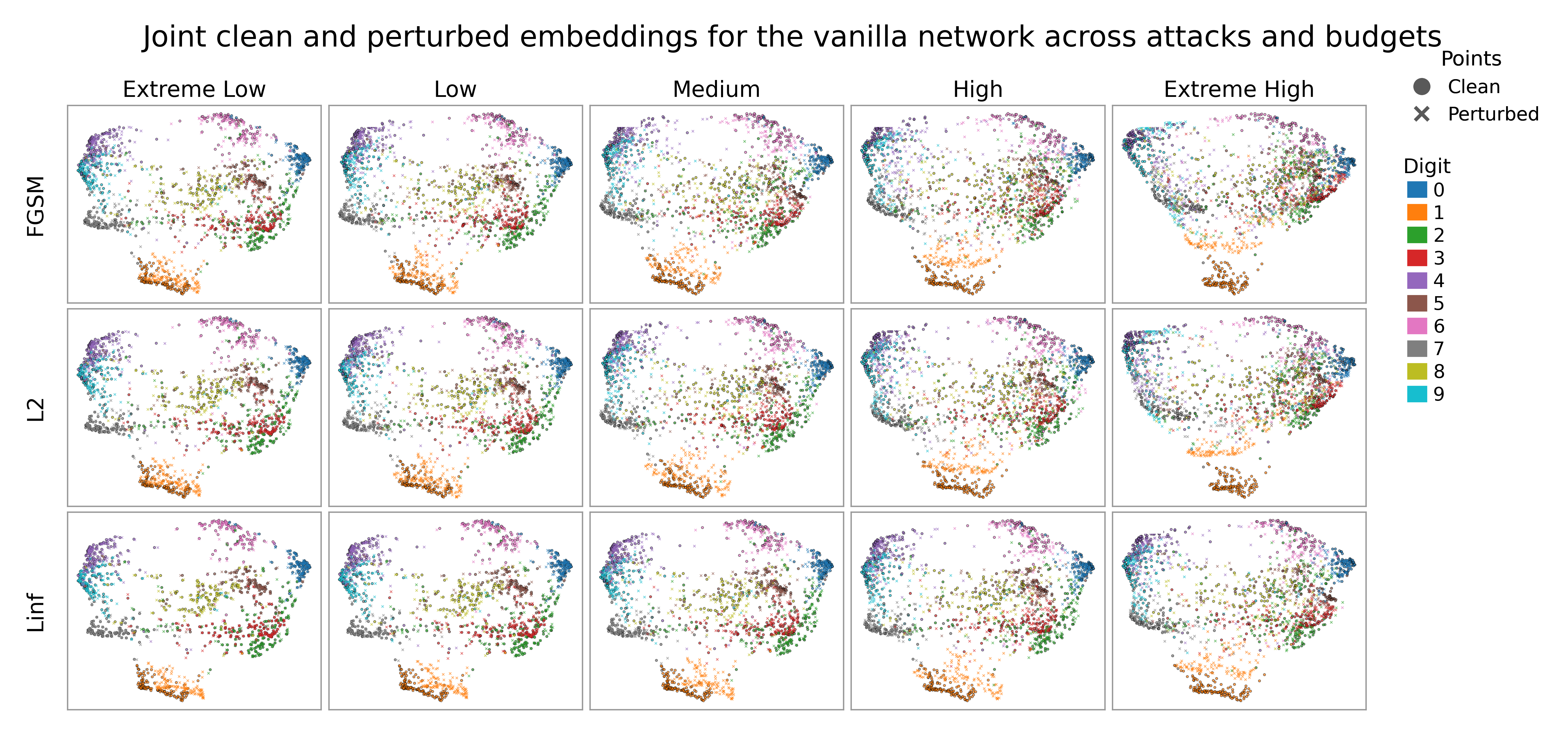}
\end{figure}

\begin{figure}[h]
  \centering
  \caption{Embeddings under the FGSM attack. Joint M-PHATE embeddings of clean (circles) and perturbed (crosses) penultimate-layer representations of 1000 MNIST test images, colored by digit class, across perturbation budgets (rows) and training regimes (columns).}
  \label{fig:fgsm_network_joint}
  \includegraphics[width=1.0\linewidth]{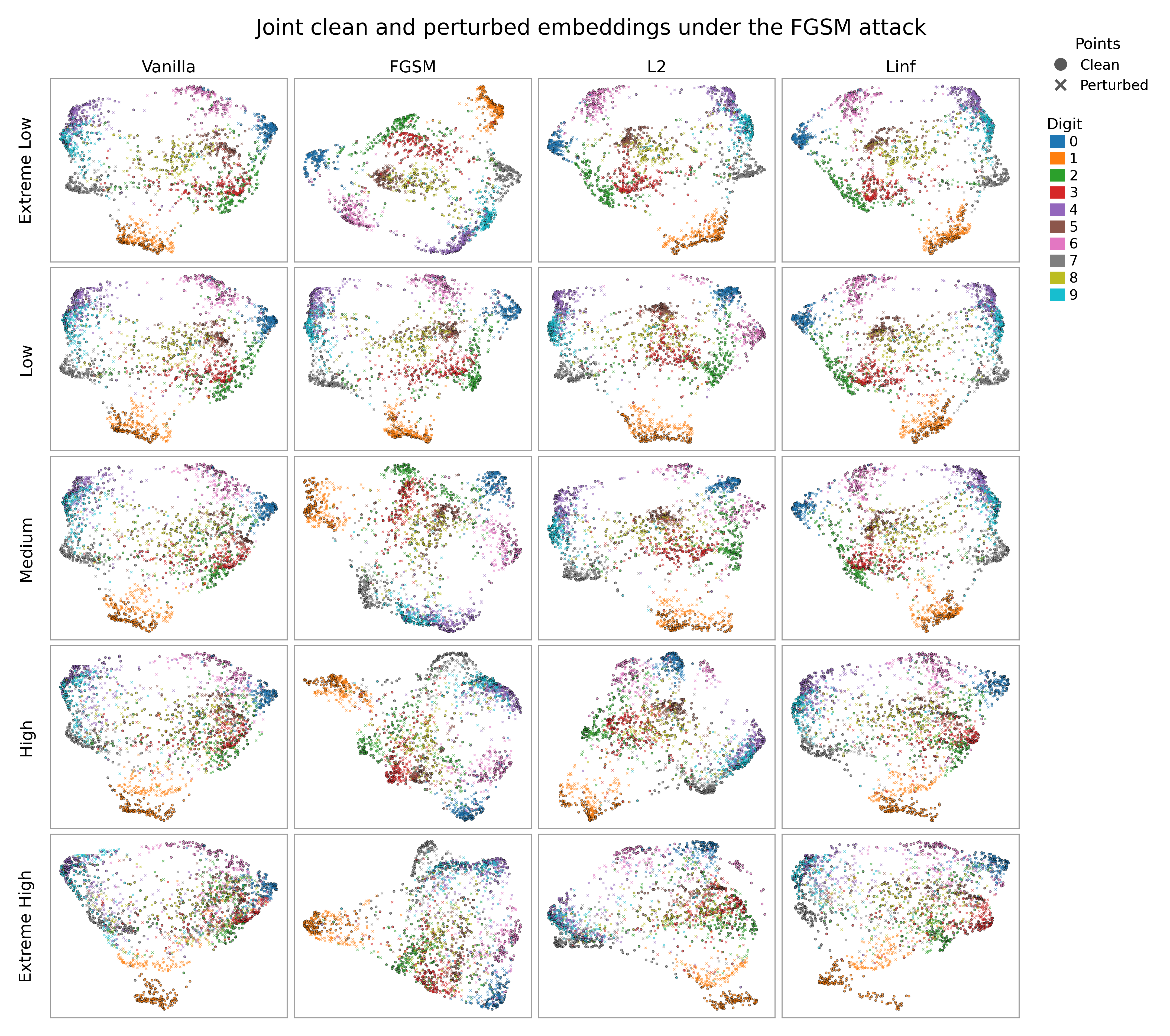}
\end{figure}

\begin{figure}[h]
  \centering
  \caption{Embeddings under the $l_2$ attack. Joint M-PHATE embeddings of clean (circles) and perturbed (crosses) penultimate-layer representations of 1000 MNIST test images, colored by digit class, across perturbation budgets (rows) and training regimes (columns).}
  \label{fig:l2_network_joint}
  \includegraphics[width=1.0\linewidth]{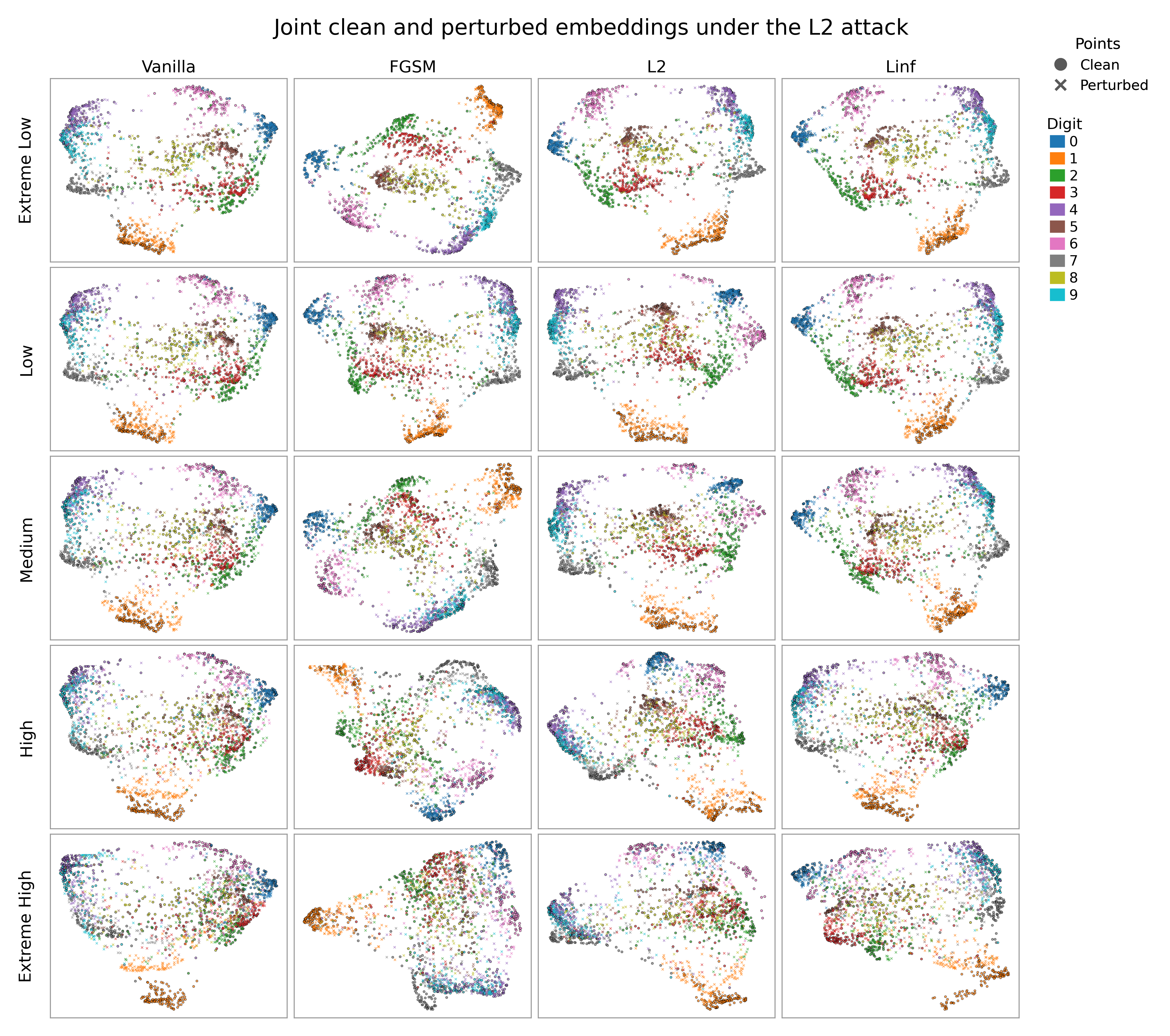}
\end{figure}

\begin{figure}[h]
  \centering
  \caption{Embeddings under the $l_{\infty}$ attack. Joint M-PHATE embeddings of clean (circles) and perturbed (crosses) penultimate-layer representations of 1000 MNIST test images, colored by digit class, across perturbation budgets (rows) and training regimes (columns).}
  \label{fig:linf_network_joint}
  \includegraphics[width=1.0\linewidth]{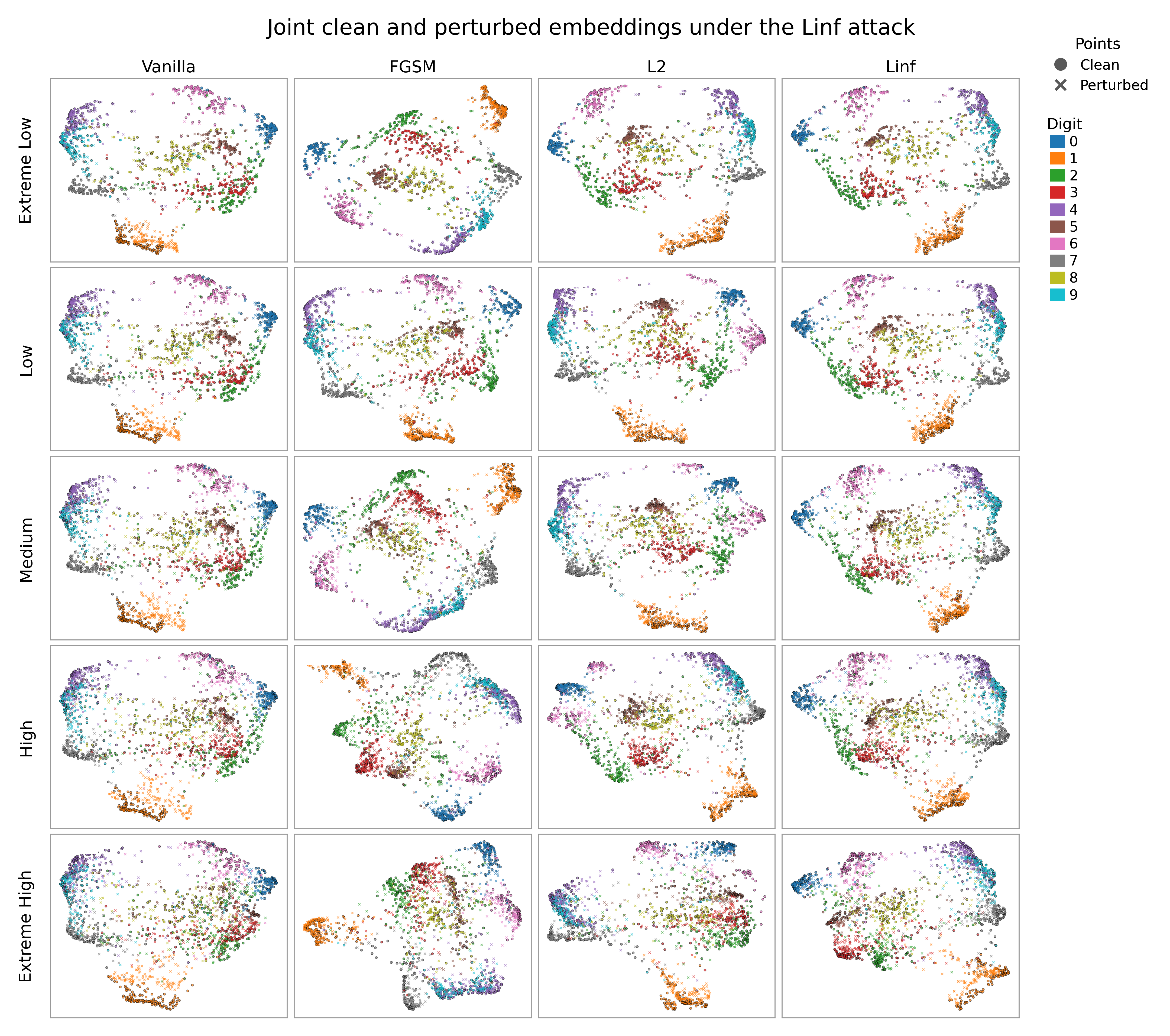}
\end{figure}

Figures \ref{fig:vanilla_network_joint}--\ref{fig:linf_network_joint} show the 2D M-PHATE embeddings of both clean and perturbed data points as penultimate layer representations ($h_2$) from each respective network, created after training has concluded.
In Figure \ref{fig:vanilla_network_joint}, within rows as perturbation budgets increase, so does the shift in perturbed point clusters from clean points and this is especially noticeable with classes for digits 0, 2, and 6.  
The $l_\infty$ attack (Figure \ref{fig:linf_network_joint}) also produces the least visible separation at every budget, compared to the FGSM and $l_2$ attack (Figures \ref{fig:fgsm_network_joint} and \ref{fig:l2_network_joint}).
The embeddings provided are best examined for qualitative patterns, as precise magnitudes and variations across seeds are given in Tables~\ref{tab:geom-raw} and~\ref{tab:geom-phate}.

\section{Computing resources}
\label{app:compute}

All experiments were run on a single machine: one NVIDIA Tesla V100 (16\,GB), a 16-core
x86\_64 CPU, and 64\,GB of system memory.

Each of the five budget sets is one self-contained run covering 3 seeds $\times$ 4 networks: 12 networks trained for 50 epochs. 
A run took roughly 20 minutes, and the five sets together took 101.9 minutes of wall-clock time. 
Network training is a minority of that cost: 26.2 minutes in total, against 35.0 minutes for the embeddings and 39.4 minutes for the permutation tests (0.9 minutes for the kNN and distance computations). 
Embedding, not training, dominates this pipeline, and the cost depends strongly on the number of slices. Averaged over exploratory runs, a 50-slice training-dynamics embedding takes $489 \pm 60$\,s, while a 2-slice clean-versus-perturbed embedding takes $124 \pm 30$\,s.

\clearpage

\end{document}